\documentclass[10pt,twocolumn,letterpaper]{article}

\usepackage{cvpr}              

\usepackage{graphicx}
\usepackage{amsmath}
\usepackage{amssymb}
\usepackage{booktabs}
\usepackage{times}
\usepackage{epsfig}
\usepackage{graphicx}
\usepackage{amsmath}
\usepackage{amssymb}
\usepackage{makecell}
\usepackage{multirow}
\usepackage{booktabs}
\usepackage{marvosym}
\usepackage{pifont}

\usepackage[pagebackref,breaklinks,colorlinks]{hyperref}

\usepackage[capitalize]{cleveref}
\crefname{section}{Sec.}{Secs.}
\Crefname{section}{Section}{Sections}
\Crefname{table}{Table}{Tables}
\crefname{table}{Tab.}{Tabs.}

\def\cvprPaperID{} 

\begin{document}

\title{ES6D: A Computation Efficient and Symmetry-Aware 6D Pose Regression Framework}


\author{\textbf{$\rm Ningkai \; Mo$} $^{1} \footnotemark[1]$ \\
\and
\textbf{$\rm Wanshui \; Gan$} $^{1, 2} \footnotemark[1]$ \\
\and
\textbf{$\rm Naoto \; Yokoya$} $^{2, 3}$ \\
\and
\textbf{$\rm Shifeng \; Chen$} $^{1} \footnotemark[2]$ \\
\and
$\rm ^1 ShenZhen \; Key \; Lab \; of \; Computer \; Vision \; and \; Pattern \; Recognition,$ \\
$ \rm Shenzhen \; Institute \; of \; Advanced \; Technology,\; Chinese \; Academy \; of \; Sciences,$ \\
$\rm ^2 The \; University \; of \; Tokyo, ^3 RIKEN$ \\
${ \rm \{nk.mo19941001, wanshuigan\}@gmail.com, yokoya@k.u-tokyo.ac.jp, shifeng.chen@siat.ac.cn}$}

\maketitle
\renewcommand{\thefootnote}{\fnsymbol{footnote}}
\footnotetext[1]{The first two authors contributed equally and should be regarded as co-first authors.}
\footnotetext[2]{Corresponding author.}


\begin{abstract}
In this paper, a computation efficient regression framework is presented for estimating the 6D pose of rigid objects from a single RGB-D image, which is applicable to handling symmetric objects. This framework is designed in a simple architecture that efficiently extracts point-wise features from RGB-D data using a fully convolutional network, called XYZNet, and directly regresses the 6D pose without any post refinement. In the case of symmetric object, one object has multiple ground-truth poses, and this one-to-many relationship may lead to estimation ambiguity. In order to solve this ambiguity problem, we design a symmetry-invariant pose distance metric, called average (maximum) grouped primitives distance or A(M)GPD. The proposed A(M)GPD loss can make the regression network converge to the correct state, i.e., all minima in the A(M)GPD loss surface are mapped to the correct poses. Extensive experiments on YCB-Video and T-LESS datasets demonstrate the proposed framework's substantially superior performance in top accuracy and low computational cost. The relevant code is available in \href{https://github.com/GANWANSHUI/ES6D.git}{https://github.com/GANWANSHUI/ES6D.git}.
\end{abstract}

\begin{figure}[t!]
\centering
{
\includegraphics[width=0.47\textwidth]{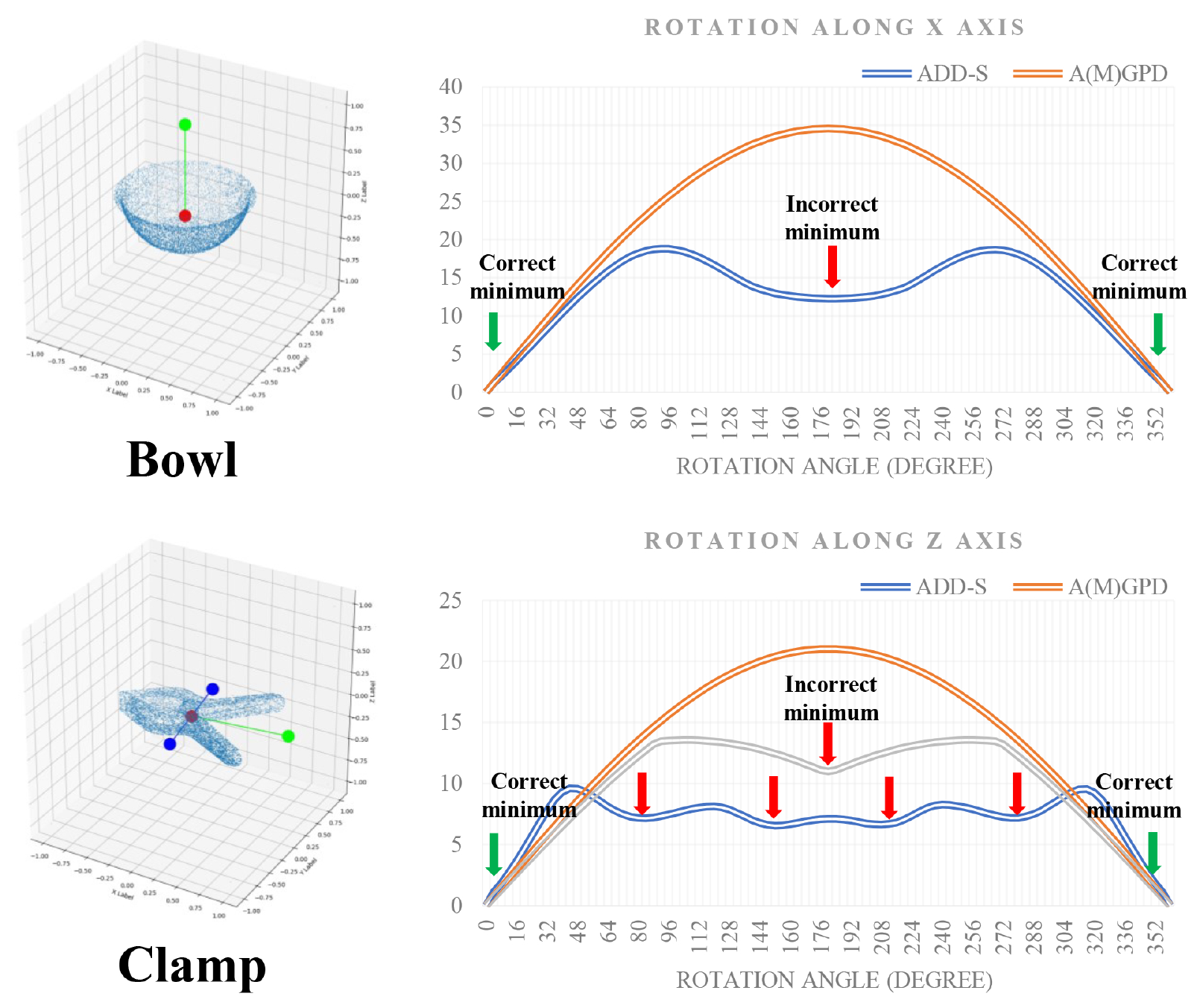}
}
\caption{\textbf{Comparison of A(M)GPD and ADD-S.} Axis $X$ shows the rotation angle of the object (from 0° to 360°). Axis $Y$ shows the calculated distance. We set the initial pose as the ground truth. As we can see, all minima are mapped to correct poses in the A(M)GPD curve and several minima point to incorrect poses in the ADD-S curve.}
\label{fig:Comparison of A(M)GPD and ADD-S}
\end{figure}

\section{Introduction}

Estimating the 6D object pose is important for real-time applications such as augmented reality (AR) \cite{marchand2015pose}, autonomous driving \cite{chen2017multi,geiger2012we}, and robotics \cite{collet2011moped,tremblay2018deep}. In recent years, methods based on the deep neural network (DNN) have gradually emerged \cite{hodavn2020bop,park2019pix2pose,zakharov2019dpod,li2019cdpn,peng2019pvnet}. 
The RGB-D-based method \cite{wang2019densefusion} fuses RGB features and point cloud features and shows exceptional robustness in handling heavy occlusion and textureless situations. However, as discussed below, regression methods \cite{xiang2017posecnn, wang2019densefusion} will fail for some symmetric objects and its computational cost is still an obstacle for real-time application. In this paper, we propose a RGB-D-based 6D pose regression framework that is more computation efficient and applicable to symmetric objects. 

\begin{figure*}
    \centering
    \includegraphics[width=1\textwidth]{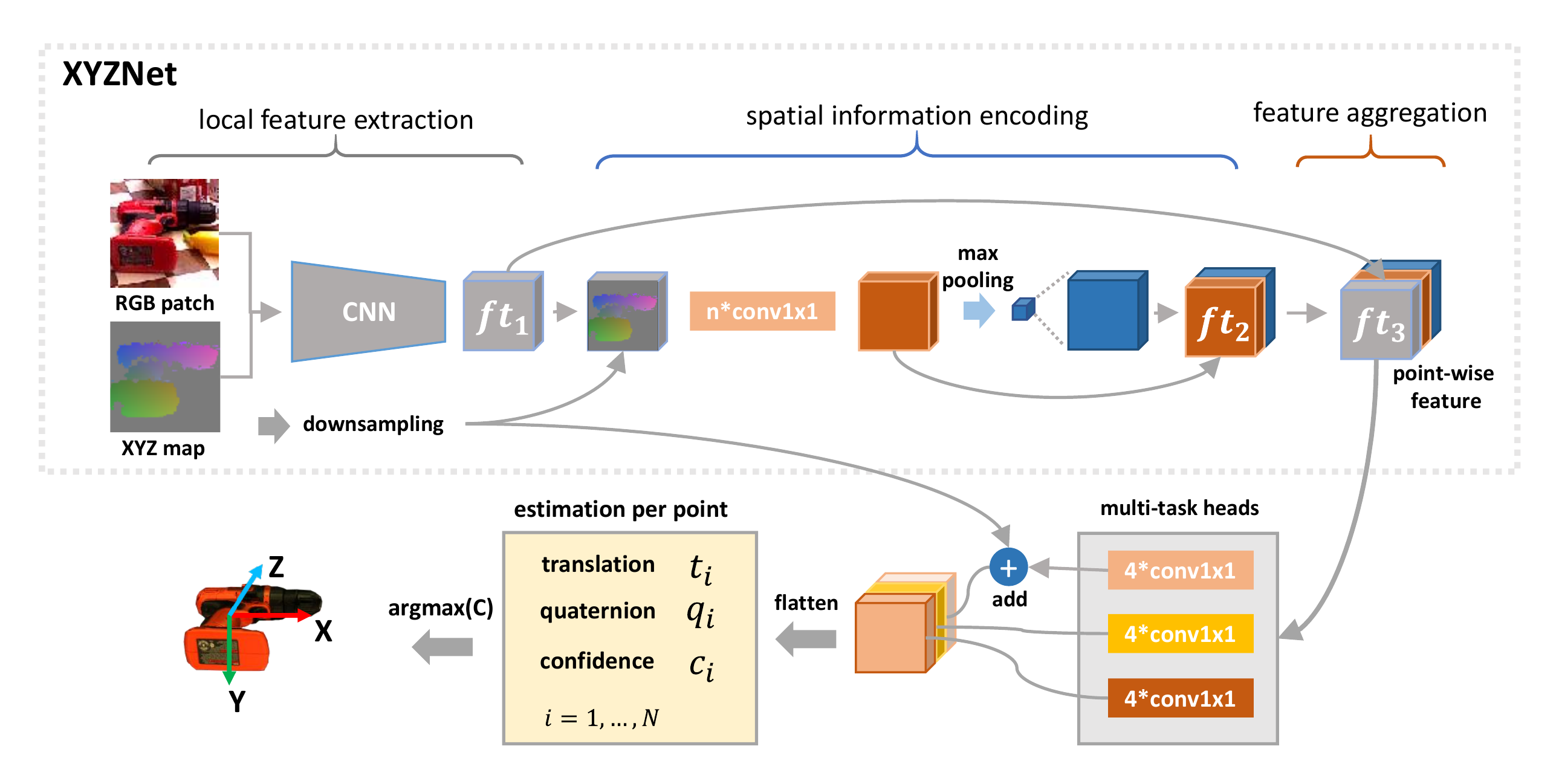}
    \caption{\textbf{Network overview.} First, the RGB-XYZ data is generated from the RGB-D image. The RGB-XYZ data is fed into a CNN module to extract local features, which encode color and geometry information. Second, the point cloud features are obtained by a PointNet-like CNN module and padded to the same size as the local features. Then, the local features and point cloud features are concatenated as the point-wise features for poses estimation. Finally, the pose with the maximum confidence is chosen as the final result.}
    \label{fig:Overview of ES6D and XYZNet}
\end{figure*}

Feature extraction from RGB-D data is a crucial part of our framework. The methods in \cite{wang2019densefusion, he2020pvn3d, hua2020rede} obtain robust features through a dense fusion network, which fuses RGB and point cloud features with an indexing operation. However, an efficient network should avoid random memory accesses \cite{2019Point}, which is the computational bottleneck of the dense fusion network in \cite{wang2019densefusion, he2020pvn3d}. For efficiency and simplicity, a fully convolutional feature extraction network, named XYZNet, is proposed in this paper. XYZNet is much more efficient than the heterogeneous structure in \cite{wang2019densefusion} and \cite{he2020pvn3d}. The depth image is converted to the XYZ map, which is strictly aligned with the RGB image, as shown in Figure \ref{fig:Overview of ES6D and XYZNet}. Therefore, the local features from RGB and the point cloud can be simultaneously extracted with a 2D convolutional kernel. Unlike the RGB-D-based method in \cite{li2018unified}, the XYZ map is propagated to the rear layer to retain the spatial information of the local features. Then, a CNN-based PointNet \cite{qi2017pointnet} module is utilized to encode the point cloud with local features. Finally, the different modality features are aggregated. The experimental results reveal the superiority of the proposed XYZNet.

In addition, learning-based manners easily fail toward the symmetric object. To explain this problem, we model the network training of 6D pose estimation as minimizing the following loss:
\begin{equation}
\setlength{\abovedisplayskip}{3pt}
\setlength{\belowdisplayskip}{3pt}
l=\operatorname{loss}(p, \hat{p})=\operatorname{loss}(N(I, w), \hat{p}),
\end{equation}
where $p$ is the estimated pose from network $N(I, w)$, $\hat{p}$ is the ground truth, $I$ represents the input image, and $w$ denotes the parameters of the network. The essence of the training is constantly adjusting the parameters of the network to the direction of the gradient of $\operatorname{loss}(p, \hat{p})$. Finally, the network will converge to the global or local minimum in the loss landscape. A symmetric object $O$ has several ground truths $S(O)=\left\{\hat{p}_{1}, \hat{p}_{2}, \ldots, \hat{p}_{k}\right\}$, which are called proper symmetries of the object $O$ \cite{Br2018Defining}. Typically, when using L1 loss to train the neural network of object $O$, it would converge to the state that predicts the average of $S(O)$, which is mapped to the minimum of the L1 loss surface. However, the average of $S(O)$ is meaningless. 

To avoid this problem, the loss function should satisfy two requirements: (1) all minima in the loss surface are mapped to the correct poses; and (2) the loss function is continuous, as Deep Networks can only approximate the continuous functions \cite{2017Universal, 1991Approximation}. \text{ADD-S} is widely used as the loss in prior regression frameworks \cite{xiang2017posecnn,wang2019densefusion,xu2019w,2020PAM} to handle symmetries. The \text{ADD-S} loss is always continuous but does not satisfy requirement (1) in some cases. As shown in Figure \ref{fig:Comparison of A(M)GPD and ADD-S}, several local minima in the \text{ADD-S} landscape are mapped to incorrect poses because of the particular shape of the objects. The motivation of our solution is to design a novel pose distance metric that is in the 3D metric space (meter, for instance) like \text{ADD-S}, and satisfies requirements (1) and (2). To this end, we introduce a novel shape representation for arbitrary objects named grouped primitives (GP). The GP is only associated with the proper symmetries $S(O)$ and ignores the details of the shape. Then, we divide symmetric objects into five categories and give the corresponding distance metric called the average (maximum) grouped primitives distance, or A(M)GPD. For typical symmetric objects, the validity of A(M)GPD is verified by a numerical and visualization method.

We evaluate the proposed framework on the YCB-Video \cite{xiang2017posecnn} and T-LESS datasets \cite{TLESS} and demonstrate its superiority by taking into account the trade-off between speed and accuracy. In particular, the experiment result on the T-LESS dataset illustrates the effectiveness of the proposed A(M)GPD loss on handling the symmetric object. 
In summary, the main contributions of this work are as follows.

\begin{itemize}
 \item We propose a novel feature extraction network XYZNet for the RGB-D data, which is suitable for pose estimation with low computational cost and superior performance.
 \item The compact shape representation GP and the distance metric A(M)GPD are introduced to handle symmetries. The loss based on A(M)GPD can constrain the regression network to converge to the correct state.
 \item A numerical simulation and visualization method is carried out to analyze the validity of the A(M)GPD loss. This analytical method is applicable to other frameworks in 6D pose estimation.
 \item The framework ES6D is proposed by using XYZNet and the A(M)GPD loss and achieves competitive performance on the YCB-Video and T-LESS datasets.

 \end{itemize}

\section{Related Work}

\subsection{Pose estimation from RGB-D data}
 To make good use of the texture and geometry information of the RGB-D data, works in \cite{wang2019densefusion,he2020pvn3d,hua2020rede,FFF6D} leverage a dense fusion network to fuse RGB and point cloud features by the indexing operation. However, the indexing operation is inefficient due to random memory access. The algorithm in \cite{li2018unified} relates work to our network, as it also tries to extract RGB and point cloud features simultaneously with 2D convolutional kernels. However, the geometric information of the point cloud is discarded during the convolution operation, which results in lower estimation accuracy. Unlike the above methods, our framework introduces a fully convolution network, XYZNet, to obtain the point-wise features, from which poses will be regressed. Moreover, none of \cite{wang2019densefusion,he2020pvn3d,hua2020rede,li2018unified,FFF6D} can handle symmetries.

\subsection{Handling symmetries in pose estimation}
A symmetric object with different poses can have an identical appearance, which leads to ambiguity as described in \cite{pitteri2019object}. To solve this problem, the methods in \cite{rad2017bb8, pitteri2019object} limit the range of rotation in the training phase and use an additional classifier to identify the range of a rotation in the testing phase. The methods in \cite{park2019pix2pose,wang2019normalized} calculate the average distance of the corresponding pixels of all the proper symmetries $S(O)$, and choose the minimum as the final loss. The object is represented by compact surface fragments in \cite{hodan2020epos}, which enable the symmetries to be handled in a systematic manner. The regression methods \cite{xiang2017posecnn,wang2019densefusion} avoid ambiguity by utilizing \text{ADD-S} as the loss in the training stage. The \text{ADD-S}, however, is not suitable to some symmetric objects, e.g., the bowl and large clamp in the YCB-Video dataset, as shown in Figure \ref{fig:Comparison of A(M)GPD and ADD-S}. Three ambiguity-invariant pose distance metrics ACPD, MCPD, and VSD proposed in \cite{hodavn2016evaluation} evaluate the error between the estimated pose and ground-truth pose. However, whether the surface of these metrics have incorrect minima has not been identified. Compared with the above methods, our A(M)GPD loss satisfies the following two properties at the same time: (1) all minima in the loss surface are mapped to the correct poses; and (2) the loss function is continuous. 
\section{The Proposed Method}
\subsection{Overview}

The aim of this paper is to detect rigid objects and estimate the corresponding rotations $R \in {S O}(3)$ and translations $\boldsymbol{t} \in \mathbb{R}^{3}$ in the camera coordinate system from an RGB-D image. A two-stage scheme is proposed as below. 

In the first stage, the segmentation network of PoseCNN \cite{xiang2017posecnn} is utilized to obtain the mask and bounding box of the target object. Each mask and RGB-D image patch cropped by the bounding box is transmitted to the second stage.

In the second stage, a real-time framework, called ES6D, is proposed to estimate the pose. The pipeline of this framework is illustrated in Figure \ref{fig:Overview of ES6D and XYZNet}. First, the masked depth pixels are transformed into the XYZ map after normalization. Second, the XYZNet extracts the point-wise features from the concatenation of the RGB patch and XYZ map. Then, three convolution heads are utilized to predict the point-wise translation offsets, quaternions, and confidences. Finally, the pose with the maximum confidence is chosen as the final result.

\subsection{Point-wise feature extraction}

It has been verified that the point-wise feature from RGB-D data is more effective and robust than the feature from the RGB image for 6D pose estimation \cite{wang2019densefusion,he2020pvn3d}. The state-of-the-art method PVN3D \cite{he2020pvn3d} adopts a heterogeneous structure that obtains the point cloud features by PointNet++ \cite{qi2017pointnet++}, and then concatenates the point cloud features with the RGB features through the indexing operation. PointNet++ extracts the local features by a series of set abstraction layers (SAL) that groups the point cloud in a pre-defined search radius. However, dealing with the massive point cloud is time-consuming, and the representation ability would decrease if we cut down the set abstraction layer. One trait of the 2D convolution operation is grouping neighboring information to extract local features. Therefore, the proposed XYZNet intends to simultaneously extract the local features by doing the 2D convolution operation on the RGB-XYZ image.

First, the masked depth pixels are transformed into the point cloud $\mathcal{P}=\left\{\left(x_{i}, y_{i}, z_{i}\right)\right\}_{i=1}^{N}$, and then the points $P$ are translated and scaled to $[-1,1]$ with the center of points $\boldsymbol{p}_{c}=\operatorname{mean}(\mathcal{P})$ and a scale factor $\gamma$. The normalized points  are denoted as $\dot{\mathcal{P}}=\left\{\left(\dot{x}_{i}, \dot{y}_{i}, \dot{z}_{i}\right)\right\}_{i=1}^{N}$ and formatted as an XYZ map. The strictly aligned RGB-XYZ data can be obtained by concatenating the XYZ map with the corresponding RGB patch. The method in \cite{li2018unified} also adopts the 2D convolution network to extract point cloud features from the XYZ map, but the performance is far worse than the heterogeneous structure methods \cite{wang2019densefusion,he2020pvn3d}. The main reason for this is that the spatial information of the point cloud would be discarded when using the 2D convolution operation on the XYZ map. We design the XYZNet based on the above observations, as illustrated in Figure \ref{fig:Overview of ES6D and XYZNet}.

The XYZNet consists of three parts. \textbf{(1) Local feature extraction module}. 2D convolution layers are used to learn the local features. The different convolution kernel sizes and the downsample rates are set to enlarge the receptive field. \textbf{(2) Spatial information encoding module}. The main function of this module is to extract the point cloud features. The module concatenates the local features with the XYZ map to regain the spatial structure and utilizes the $1\times1$ convolution to encode the local feature and coordinate of each point. Then, the global feature is obtained by max-pooling and concatenated to each point feature to provide a global context. \textbf{(3) Feature aggregation}. The local features and point cloud features are concatenated as the point-wise features. The fusion of the two modalities makes pose estimation robust against less texture and heavy occlusion.

\subsection{6D pose regression}

After the XYZNet is completed, the set of point-wise features $F=\left\{\boldsymbol{f}_{i}\right\}_{i=1}^{N}$, $\boldsymbol{f}_{i} \in \mathbb{R}^{d}$, are obtained. In this subsection, we describe how to exploit the point-wise feature $\boldsymbol{f}_{i}$ and the corresponding visible point $\boldsymbol{\dot{p}}_{i} \in \dot{\mathcal{P}}$ to estimate the rotation $R_{i} \in {S O}(3)$ and translation $\boldsymbol{t}_{i} \in \mathbb{R}^{3}$. As shown in Figure \ref{fig:Overview of ES6D and XYZNet}, three $1\times1$ convolution heads $\left(\mathcal{B}_{\mathcal{T}}, \mathcal{B}_{\mathcal{Q}},\mathcal{B}_{C}\right)$ are adopted to regress the translation offset $\left(\Delta \boldsymbol{\dot{t}}_{i} \in \mathbb{R}^{3}\right)$, quaternion $\left(\boldsymbol{q}_{i} \in \mathbb{R}^{4},\left\|\boldsymbol{q}_{i}\right\|=1\right)$ and confidence $\left(c_{i} \in[0,1]\right)$.

\textbf{3D translation regression} Regarding the origin of the normalized object coordinate system as a virtual keypoint, the translation $\boldsymbol{t}_{i}$ can be obtained by calculating the offset $\Delta \boldsymbol{\dot{t}}_{i}$ between the visible point $\boldsymbol{\dot{p}}_{i}$ and the origin. The equation could be given as: 

\begin{equation}
\Delta \boldsymbol{\dot{t}}_{i}=\mathcal{B}_{\mathcal{T}}\left(\boldsymbol{f}_{i}\right),
\end{equation}
\begin{equation}
\boldsymbol{t}_{i}=\frac{\left(\boldsymbol{\dot{p}}_{i}+\Delta \boldsymbol{\dot{t}}_{i}\right)}{\gamma} +\boldsymbol{p}_{c},
\end{equation}
where the offset of the visible point $\boldsymbol{\dot{p}}_{i}$ is distributed in a specific sphere. This regression function gets a smaller output space than directly regressing the object translation \cite{gao20206d}. 

\textbf{3D rotation regression} We exploit the quaternion as rotation representation following \cite{xiang2017posecnn, wang2019densefusion}. We get the rotation matrix as follow:
\begin{equation}
R_{i}= Quaternion\_matrix\left({Norm}\left(\mathcal{B}_{Q}\left(\boldsymbol{f}_{i}\right)\right)\right),
\end{equation}

\begin{equation}
    {Norm}(\boldsymbol{q}_{i})=\frac{\boldsymbol{q}_{i}}{\|\boldsymbol{q}_{i}\|},
\end{equation}
where $Quaternion\_matrix(\cdot)$ denotes the function that transforms the quaternion into the rotation matrix \cite{sarabandi2019survey}. 

\textbf{Confidence regression} To identify the best regression result, we set a confidence estimation head to evaluate each feature's confidence $c_{i}$. The equation is given as: 
\begin{equation}
    c_{i}={Sigmoid}\left(\mathcal{B}_{C}\left(\boldsymbol{f}_{i}\right)\right).
\end{equation}
We train the confidence branch $\mathcal{B}_{C}$ with the self-supervision approach that is mentioned in \cite{wang2019densefusion}.

\subsection{Symmetry-aware loss}

\begin{figure*}
    \centering
    \includegraphics[width=1\textwidth]{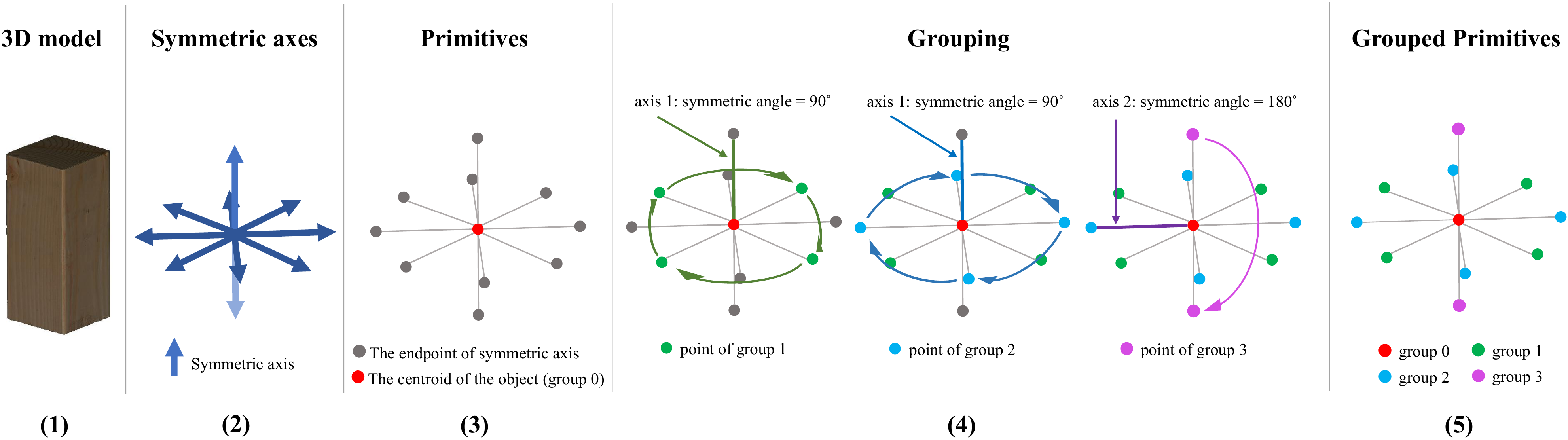}
    \caption{\textbf{The pipeline of the GP construction.}}
    \label{fig:GP construction}
\end{figure*}


The existed symmetry-invariant distance metric depends on
the 3D shape of the object, such as ADD-S, ACPD, MCPD, VSD \cite{hodavn2016evaluation, wang2019densefusion}. However, unique shape and point-pair mismatch are the causes of incorrect minima. Besides, objects, in reality, have various
shapes and we cannot guarantee these metrics are valid for
every shape. Therefore, we designed grouped primitives, GP, that abstract objects of the same category into several points to avoid the
uncertainty caused by the shape. Furthermore, we divide these
points into groups and calculate the distance between closest
points in the same group, according to Eq. \ref{AGPD} and \ref{MGPD},
which avoids point-pair mismatch.


\textbf{Grouped primitives} We illustrate the
pipeline of the GP construction in Figure \ref{fig:GP construction}. Having the 3D
model of the specific object, we could calculate all symmetry
axes according to Eq. \ref{AO} and \ref{MAO}. The primitives for grouping
are composed of the endpoint of the symmetry axis and the
object centroid. Specifically, the following three steps are required.

\textbf{Step 1} The basic properties of the symmetry axis-angle are defined and explained. The appearance of the object $O$ looks the same after a rotation around axis $\boldsymbol{e}=(e_x, e_y, e_z)$ by angle $\theta$. Thus, the axis $\boldsymbol{e}$ is a symmetry axis of $O$. The axis $\boldsymbol{e}$ and the angle $\theta$ compose a symmetry axis-angle $\boldsymbol{a}$ , which is defined as:
\begin{equation}
    \boldsymbol{a}=(\boldsymbol{e}, \theta), \quad \|\boldsymbol{e}\|=1 \wedge \theta \in\{2 \pi / i\}_{i=2}^{M}.
\end{equation}
It is important to note that $2\pi$ must be an integer multiple of the angle $\theta$ of symmetry \cite{weyl2015symmetry}, and the order of $\boldsymbol{a}$ can be defined as:
\begin{equation}
\setlength{\abovedisplayskip}{3pt}
    |\boldsymbol{a}|=2 \pi / \theta(\boldsymbol{a}).
\end{equation}

\begin{figure*}
    \centering
    \includegraphics[width=1\textwidth]{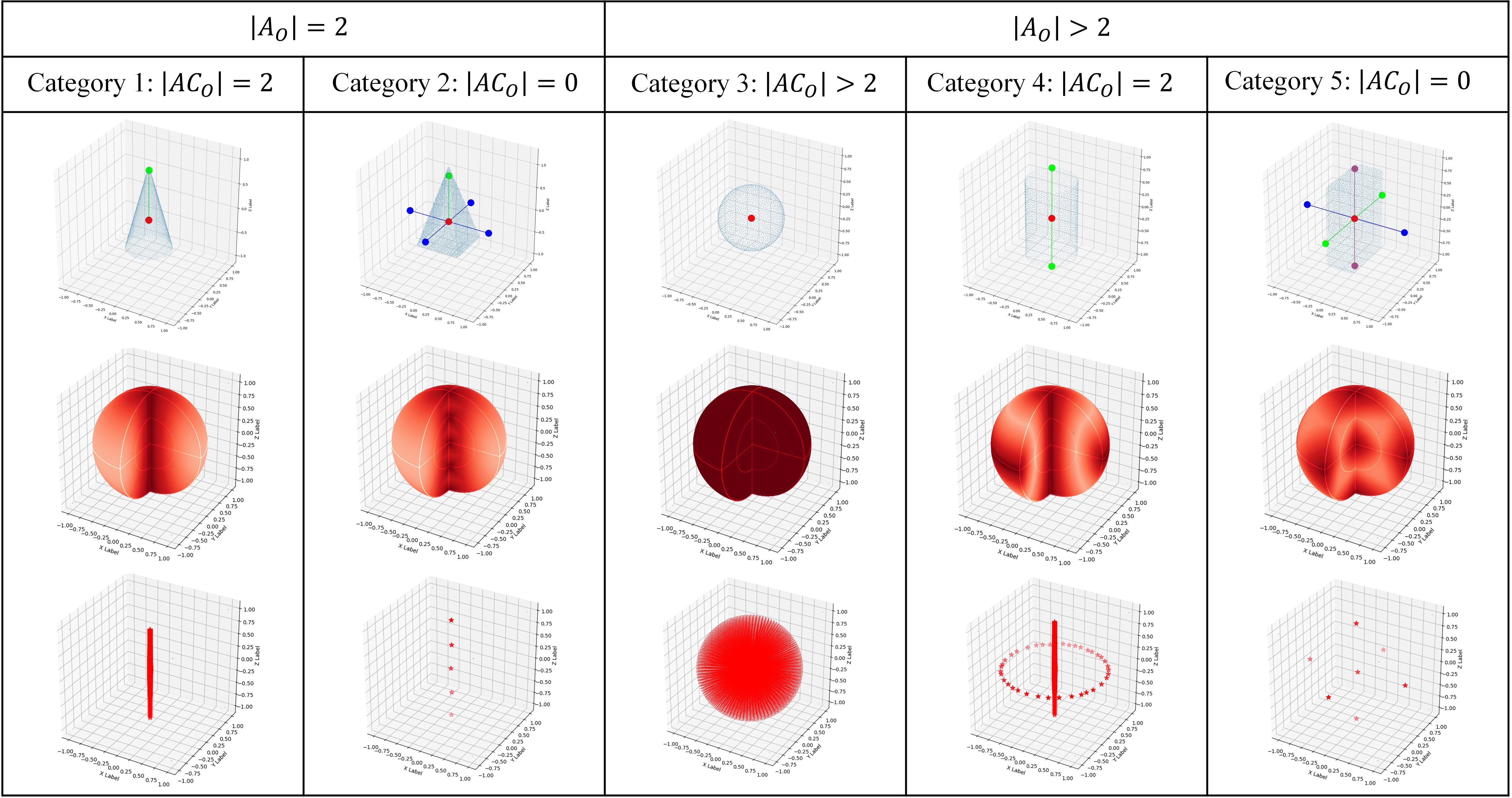}
    \caption{\textbf{Grouped primitives and the visualization of A(M)GPD landscape.} Based on the size of ${A}_{O}$ and $AC_{O}$, symmetric objects can be classified into five categories. For each category, a typical toy model and its grouped primitives are presented in the first row plots. The second row shows the A(M)GPD landscape of each object in the rotation space, where the darker color represents the smaller value of A(M)GPD. The third row shows the minima in each landscape. Best viewed in color.}
    \label{fig:Grouped primitives and the visualization of A(M)GPD landscape}
\end{figure*}

The symmetry axis-angle is a redundant form. For example, a pyramid, the object of category 2 in Figure \ref{fig:Grouped primitives and the visualization of A(M)GPD landscape}, has four symmetry axis-angles: $(\boldsymbol{e}, \pi / 2)$, $(\boldsymbol{e}, \pi)$, $(-\boldsymbol{e}, \pi / 2)$, and $(-\boldsymbol{e}, \pi)$, where $\boldsymbol{e}$ is  parallel to the green line. In this case, the four symmetry axis-angles have the same meaning for this object because of the same axis $\boldsymbol{e}$. The angles of these four symmetry axis-angles must have a greatest common divisor $\pi / 2$ due to the cyclic property of rotational symmetry. \textbf{Note that, only the symmetry axis-angle, whose angle is the greatest common divisor, is used in this work}, e.g.,  $(\boldsymbol{e}, \pi / 2)$ and $(-\boldsymbol{e}, \pi / 2)$.

\textbf{Step 2} In the object coordinate system, in which the centroid of the object is used as the origin, a set of rough symmetry axis-angles of object $O$ can be obtained by using the following formula:
\begin{equation}
\hat{A}_{O}=\left\{\boldsymbol{a} | h\left(P_{O}, R(\boldsymbol{a}) P_{O}\right)<\varepsilon\right\},
\label{AO}
\end{equation}
where $h$ is the Hausdorff distance, $P_{O}$ represents the vertices of the object model, $R(\boldsymbol{a})$ is the associated rotation matrix of symmetry axis-angle $\boldsymbol{a}$, and the allowed deviation is bounded by $\varepsilon$. Then, based on symmetry axes, the Mean-Shift clustering algorithm \cite{comaniciu2002mean} is applied to simplify $\hat{A}_{O}$:
\begin{equation}
A_{O}=Mean\_Shift(\hat{A}_{O}).
\label{MAO}
\end{equation}
At this point, $A_{O}$ contains all symmetry axis-angles of the object $O$ without redundancy, where $|A_{O}|$ is the size of $A_{O}$ and is a multiple of 2 because symmetry axis-angles always come in pairs, e.g.,  $(\boldsymbol{e}, \pi / 2)$ and $(-\boldsymbol{e}, \pi / 2)$. Further, a subset $AC_{O}$ of $A_{O}$ can be obtained as:
\begin{equation}
A C_{O}=\{\boldsymbol{a}|\boldsymbol{a} \in A_{O} \wedge |\boldsymbol{a}|>\rho\},
\end{equation}
where $\rho$ is the relaxed threshold. When $|\boldsymbol{a}|>\rho$,  we consider $\boldsymbol{a}$ as a continuous symmetry axis-angle, and most of the applications are covered when $\rho$ is set as 6, including all the objects to be evaluated in the experiment section. 
According to the size of $A_{O}$ and $AC_{O}$, symmetry objects can be divided into five categories, as shown in Figure \ref{fig:Grouped primitives and the visualization of A(M)GPD landscape}.


\textbf{Step 3} As illustrated in Figure \ref{fig:GP construction}, if the primitive A could overlap with primitive B after a specific angle around the axis of symmetry, we regard primitive A and B lie in the same group. The grouped primitives are denoted as $G=\left\{g_{i}\right\}_{i=0}^{K}$, where $K$ is the size of $G$. The details of grouping principle are showed in the supplementary material.

 \textbf{Pose distance metric} Based on the GP, the pose distance metric A(M)GPD is designed. The A(M)GPD contains two functions, the first of which is average grouped primitives distance (AGPD):
\begin{equation}
\setlength{\abovedisplayskip}{3pt}
     A G P D=\operatorname{mean}_{g_{i} \in G} \operatorname{mean}_{p_{j} \in g_{i}} \operatorname{min} _{p_{k} \in g_{i}, k \neq j}\left\|\hat{p}_{j}-\dot{p}_{k}\right\|,
     \label{AGPD}
 \end{equation}
 where $\hat{p}=\widehat{T} p, \dot{p}=\dot{T} p, p \in g(G)$, and  $\widehat{T},\dot{T} \in {SE}(3)$. AGPD is used to measure the distance of two poses of object $O$, when $O$ is one of the  symmetry categories $\left\{ 1, 3, 4, 5 \right\}$ or the asymmetric object.
 
The category 2 is different from the others. It has only one pair of symmetry axes, which have a finite order. This property leads to an incorrect minimum in the rotation space if AGPD is used as the loss, as illustrated in the second row in Figure \ref{fig:Comparison of A(M)GPD and ADD-S}. To solve this problem, the second function maximum grouped primitives distance (MGPD) is introduced:
 \begin{equation}
\setlength{\abovedisplayskip}{3pt}
     M G P D=\operatorname{max}_{g_{i} \in G} \operatorname{max}_{p_{j} \in g_{i}} \operatorname{min} _{p_{k} \in g_{i}k \neq j}\left\|\hat{p}_{j}-\dot{p}_{k}\right\|.
     \label{MGPD}
 \end{equation}

 \textbf{Loss for regression training} The total loss of our regression framework is similar to the loss in \cite{wang2019densefusion}, where the difference is that A(M)GPD is used to calculate the error between prediction and ground truth instead of ADD(S).

\begin{table*}[t]
\centering
\scalebox{0.78}{
\begin{tabular}{l|c|c|c|c|c|c|c|c|c|c|c|c}

\hline
 & \multicolumn{2}{c|}{ } & \multicolumn{6}{c|}{With PoseCNN segment mask} & \multicolumn{4}{c}{With GT segment mask}
 \tabularnewline

\hline
&  \multicolumn{2}{c|}{FFB6D}&  \multicolumn{2}{c|}{DenseFusion }&  \multicolumn{2}{c|}{DenseFusion} &  \multicolumn{2}{c|}{\multirow{2}*{ES6D}} &   \multicolumn{2}{c|}{PVN3D \cite{he2020pvn3d}} &  \multicolumn{2}{c}{\multirow{2}*{ES6D}} \\

&  \multicolumn{2}{c|}{\cite{FFF6D}}& \multicolumn{2}{c|}{(per-pixel) \cite{wang2019densefusion}}&  \multicolumn{2}{c|}{(iterative) \cite{wang2019densefusion}} &  \multicolumn{2}{c|}{~}&   \multicolumn{2}{c|}{(post process)} &  \multicolumn{2}{c}{~} 
\tabularnewline

\hline

&ADD-S &ADD(S)&ADD-S &ADD(S) &ADD-S &ADD(S) &ADD-S &ADD(S) &ADD-S &ADD(S) &ADD-S & ADD(S) 
\tabularnewline
\hline

bowl & 96.3 & 96.3 & 86.0 & 86.0 & 89.5 & 89.5 & 96.4 & 96.4 & 88.7 & 88.7 & 96.8 & 96.8 

\tabularnewline

wood\_block & 92.6 & 92.6 & 89.5 & 89.5 & 92.8  & 92.8 & 94.4  & 94.4 & 91.5 &  91.5 & 96.0  & 96.0 

\tabularnewline

large\_clamp & 96.8 & 96.8 & 71.5 & 71.5 & 72.5 & 72.5  & 61.0  & 61.0 & 94.4 & 94.4 & 97.5 & 97.5

\tabularnewline

extra\_large\_clamp & 96.0 & 96.0  & 70.2 & 70.2 &  69.9 & 69.9 & 59.6 & 59.6 & 91.1 & 91.1 & 96.8 & 96.8

\tabularnewline

foam\_brick & 97.3 &  97.3 & 92.2 & 92.2 & 92.0 & 92.0  & 96.6 & 96.6 & 96.8 & 96.8 & 96.9 & 96.9 

\tabularnewline
\hline

ALL & \textbf{96.6} & \textbf{92.7} & 91.2 & 82.9 & 93.2 & 86.1 & \textbf{93.6}  & \textbf{89.0} & 95.7 & 91.9 &  \textbf{97.1} & \textbf{93.2}

\tabularnewline
\hline

\end{tabular}
}

\caption{Comparison of 6D pose (ADD-S, ADD(S)) on the YCB-Video dataset \cite{xiang2017posecnn}. The listed objects are symmetric. More detail could be found in the supplementary material.}
\label{t:YCB-Video dataset}
\end{table*}
\begin{table*}[t]
\centering
\scalebox{0.78}{
\begin{tabular}{l|c|c|c|c|c|c|c|c|c}

\hline

& Pose Est. & $\textit{e}_{ADI}$  \rm (VIVO) & $\textit{e}_{VSD}$  \rm (VIVO) & $\textit{e}_{ADI}$  \rm (SISO) & $\textit{e}_{VSD}$  \rm (SISO) & ADD(S) & A(M)GPD &Training data & Time (s) 
\tabularnewline
\hline

PointNet++ \cite{qi2017pointnet++}  & D & 0.74  & 0.50 & 0.78 & 0.54  & -- &  -- & 37K & 0.4

\tabularnewline

PPFNet \cite{deng2018ppfnet} & D &  0.76 & 0.44 & 0.79  & 0.49  & -- &  -- & 37K & 0.4

\tabularnewline

StablePose \cite{shi2021stablepose} & D & \textbf{0.86}  & 0.69 & \textbf{0.88} & 0.73 & -- &  -- & 37K & 0.4

\tabularnewline
\hline

Pix2Pose \cite{park2019pix2pose} & RGBD & --  & -- & -- & 0.30  & -- & --  & 37K & 0.6

\tabularnewline

CosyPose \cite{labbe2020cosypose} & RGBD  & 0.68  & 0.63 & 0.75  &  0.64 &  -- & --  & 1M  & 1.1

\tabularnewline
\hline

ES6D (ADD(S)) & RGBD & 0.79 &0.68  & 0.80 & 0.69  &   93.08& 55.99  & 1M & \textbf{0.07}

\tabularnewline

ES6D (A(M)GPD) & RGBD & 0.81& \textbf{0.75} & 0.82 & \textbf{0.76}  &  \textbf{93.40} & \textbf{82.70}  & 1M & \textbf{0.07}

\tabularnewline

\hline

\end{tabular}
}

\caption{Comparison of 6D pose on the T-LESS dataset \cite{TLESS}. ES6D (ADD(S)) and ES6D (A(M)GPD) means the network is trained by the ADD(S) and A(M)GPD loss, respectively. The inference time of ES6D includes the mask segmentation cost. }
\label{t:TLESS1 dataset}
\end{table*}

 \subsection{Validation of A(M)GPD}
 
 In this subsection, a numerical and visualization method is proposed to check whether A(M)GPD meets the requirement (1) described in the introduction. In order to get a clearer view of the A(M)GPD landscape on $R \in {S O}(3)$, we first exploit the sampling technique to generate $N$ rotations
 $RC=\{R_{i}\}_{i=1} ^ N$ that are densely distributed on $R \in {S O}(3)$. Second, the identity 
matrix $I_{3\times3}$ is treated as the ground truth, and $\dot{R} \in RC$ is the prediction. The A(M)GPD of $I_{3 \times 3}$ and $\dot{R}$ can be given as $\dot{d}$,
\begin{equation}
    \dot{d}=\operatorname{A(M)GPD}(I_{3 \times 3}, \dot{R}).
\end{equation} 
Then, we visualise $\dot{d}$ with the help of the rotation vector $\boldsymbol{v}=(v_{x},v_{y},v_{z})$, in which the direction is the rotation axis and the length is the rotation angle $\theta \in [0, \pi]$. As shown in the second row plots in Figure \ref{fig:Grouped primitives and the visualization of A(M)GPD landscape}, the coordinate of $\dot{R}$ is $\boldsymbol{v}(\dot{R})$ and the color value of $\dot{R}$ is the corresponding $\dot{d}$ (the darker color represents the smaller $\dot{d}$). However, it is hard to find minima in these plots, so we further simulate the process of gradient descent by a simple algorithm. The principle of this algorithm is that $\boldsymbol{v}(\dot{R})$ constantly moves to $\boldsymbol{v}(\hat{R})$, which has the minimum $\hat{d}$ in the neighborhood of $\boldsymbol{v}(\dot{R})$ and this point will stop in a local minimum at last. We perform this principle on each $\boldsymbol{v}(\dot{R})$, and the found minima are labeled with red stars in the third row plots in Figure \ref{fig:Grouped primitives and the visualization of A(M)GPD landscape}. As we can see, all minima are mapped to the correct poses. The other objects are presented in the supplement.

\section{Experiments}
\subsection{Implementation detail}
Our approach is implemented with Pytorch. We resize the RGB patch and XYZ map into $128\times128$ before putting them into the neural network. The local feature extraction module in the XYZNet is modified from ResNet18 \cite{he2016deep}. For better performance, the grouped primitives is scaled by the object's radius. All the experiments are on an Intel (R) Xeon (R) 2.4GHz CPU with NVIDIA GTX 2080 Ti GPU. 
\begin{table*}[t]
\centering
\scalebox{0.9}{
\begin{tabular}{l|c|c|c|c|c|c|c|c|c}

\hline

\multirow{2}{*}{Method} & \multirow{2}{*}{LEF} & \multirow{2}{*}{CXYZ} & \multirow{2}{*}{SIE} & \multirow{2}{*}{FA} & A(M)GPD & YCB & Time & FLOPs  & Parameters\\

 &  &  & &  & Loss  & ADD(S) & (ms) & (G) & (M)

\tabularnewline
\hline

Unified\_like \cite{li2018unified} &  &  & & &  & 91.50 & 8.4 &  7.39 & 17.85
\tabularnewline
\hline


XYZNet\_1 & Res18  &  & &  &  & 91.86  & 5.1 & 7.90 &16.29
\tabularnewline
\hline

XYZNet\_2 & Res18 &  & $\surd$ &  &  & 92.04 & 5.7 & 9.16 & 17.52
\tabularnewline
\hline

XYZNet\_3 & Res18 & $\surd$  & $\surd$ &  & & 92.42 & 5.8 & 9.16 & 17.52
\tabularnewline
\hline

XYZNet\_4 & Res18  & $\surd$  & $\surd$ & $\surd$ &  & 93.03 & 5.9 & 10.17 & 18.51
\tabularnewline
\hline

ES6D & Res18  & $\surd$  & $\surd$ & $\surd$ & $\surd$ & \textbf{93.23} & 5.9 & 10.17 & 18.51
\tabularnewline
\hline

\end{tabular}
}
\caption{Ablation study on XYZNet. LFE: local feature extraction; CXYZ: concatenate XYZ map and local features; SIE: spatial information encoding; FA: feature aggregation. The detailed structure of each module is illustrated in Figure \ref{fig:Overview of ES6D and XYZNet}. } 
\label{t:AblationXYZNet}
\end{table*}

\subsection{Datasets}

\textbf{YCB-Video} \cite{xiang2017posecnn} is collected from 21 YCB \cite{calli2015ycb} objects including 5 symmetric objects, which is a challenging task due to its various lighting conditions, significant image noise, and occlusions. The dataset contains 92 RGB-D videos, where each video shows a subset of the 21 objects in different indoor scenes. We follow prior works and split the dataset into 80 videos for training and 2,949 keyframes from the remaining 12 videos for testing. We also use the 80,000 synthetic images released by \cite{xiang2017posecnn} in our training set.

\textbf{T-LESS} \cite{TLESS} is a challenging dataset with 27 symmetric objects and 3 asymmetric objects, which could effectively evaluate our proposed symmetric-aware method. Since the object is texture-less and has a similar appearance feature, it is much more challenging than the YCB-Video dataset. We use the mask result from \cite{shi2021stablepose} for a fair comparison.

\subsection{Metrics}

In YCB-Video dataset, following \cite{he2020pvn3d}, the area under curve (AUC) of ADD-S and ADD(S) is treated as performance metrics for comparison of peer algorithms. In addition, the ADD(S) \cite{hinterstoisser2012model} calculates the ADD distance for non-symmetric objects and ADD-S distance for symmetric objects, which is more rigorous in evaluation than ADD-S. The AUC of ADD-S and ADD(S) for the YCB-Video dataset serve as the performance metrics.


In T-LESS dataset, we report the Average Closest Point Distance (ADI) and Visible Surface Discrepancy (VSD) following the setting in \cite{shi2021stablepose}. In addition, to reveal the discrepancy of ADD(S) and A(M)GPD, we compare the AUC of ADD(S) and the proposed A(M)GPD for the ablation study with the ground truth mask because the mask from \cite{shi2021stablepose} does not offer the index to the ground truth label.

\subsection{Comparison with SOTA methods}
\textbf{YCB-Video} To ensure a fair comparison with DenseFusion \cite{wang2019densefusion}, we use the segmentation of PoseCNN for the testing results. Note that the large clamp and extra-large clamp in the dataset have the same appearance but with different sizes, which would cause a poor segmentation result. The failure cases in ES6D are shown in the supplemental material. From Table \ref{t:YCB-Video dataset}, it is observed that our method outperforms DenseFusion (iterative) by 2.9\%. FFB6D \cite{FFF6D} is better than us, which gets a better instance segmentation result by clustering after the segmentation but with an additional time cost. It is worth mentioning that no refinement and post process are used in our method, while the DenseFusion (iterative) includes the refinement and post process. In addition, we take the ground truth masks as input both in ES6D and PVN3D \cite{he2020pvn3d} for a comparison. In particular, our method outperforms PVN3D in symmetric objects by a large margin, \eg, bowl (8.1\%), wood\_block (4.5\%), large\_clamp (3.1\%), and extra\_large\_clamp (5.7\%). 

\textbf{T-LESS} Table \ref{t:TLESS1 dataset} shows the comparison of 6D pose on the T-LESS dataset \cite{TLESS}. Pix2Pose \cite{park2019pix2pose} regresses pixel-wise 3D coordinates by an auto-encoder architecture. CosyPose \cite{labbe2020cosypose} estimates the 6D pose based on the RGB image, and then does the ICP refinement with the depth image. StablePose \cite{shi2021stablepose} obtains 6D object pose by stable patch extraction and patch pose estimation. Compared with these methods, the proposed ES6D is a more simple and efficient framework. We achieve the best result in the VSD metric in both single instance of a single object (SISO) and varying number of instances of varying number objects in single-view RGBD images (VIVO). Besides, the inference time is much lower in comparison to these methods.

\subsection{Ablation study}
\textbf{XYZNet} 
We further explore the effects of the individual modules in XYZNet in Table \ref{t:AblationXYZNet}. The experiments are based on our regression framework. All methods are trained with the ADD(S) loss, except for ES6D. The experimental results demonstrate that the complete network, which comprises LEF, CXYZ, SIE, and FA, is the optimal architecture in these schemes. The Unified\_like \cite{li2018unified} structure is not satisfactory on both accuracy and inference time. Compared with XYZNet\_2, XYZNet\_3 obtains a large improvement by concatenating the XYZ map to local features, demonstrating the effectiveness of this explicit concatenation operation in practice. Furthermore, by adding the FA module, the  XYZNet\_4 yields the improvement, which illustrates the effectiveness of multimodal feature fusion (2D image and 3D point cloud).

\begin{figure}[t!]
\centering
{
\includegraphics[width=0.47\textwidth]{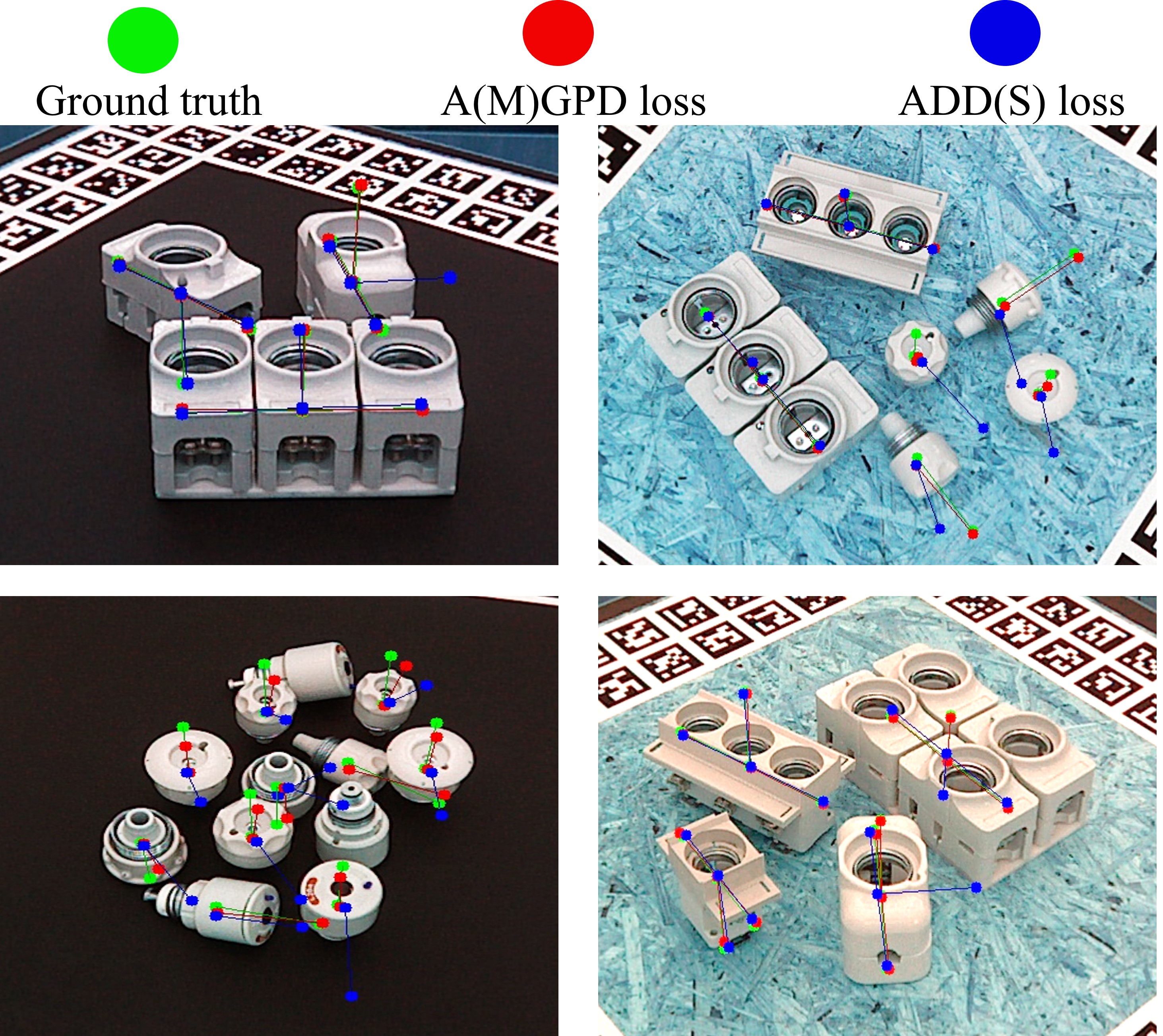}
}
\caption{\textbf{Visualization on the T-LESS dataset with different training loss.} The green, red, and blue lines represent the ground truth pose, the result from A(M)GPD loss, and the result from ADD(S) loss, respectively. }
\label{fig:Comparison of A(M)GPD and ADD-S 2}
\end{figure}

\textbf{A(M)GPD versus ADD(S)} The motivation behind developing the A(M)GPD is the problem that the ADD(S) metric is insensitive to the rotation error of symmetry objects. For the comparison between ADD(S) loss and the proposed A(M)GPD loss, we conduct the experiment on the proposed ES6D with different loss settings. From the Table \ref{t:TLESS1 dataset}, it can be seen that the AUC of ADD(S) is close but there is a large gap in the A(M)GPD metric. For a more convincing result, we visualize part of the symmetric object in Figure \ref{fig:Comparison of A(M)GPD and ADD-S 2}.  We observe that the ADD(S) loss result can have the totally reversed pose, but the ADD(S) metric can not distinguish this situation. On the other hand, we also see that the result from the A(M)GPD loss could correctly reflect this situation. By combining it with the curve illustrated in Figure \ref{fig:Comparison of A(M)GPD and ADD-S}, we can conclude that the proposed A(M)GPD loss could effectively eliminate the local minima problem in ADD-S loss during the training phase. In addition, the proposed A(M)GPD metric is much more accurate in the pose evaluation of a symmetric object. 


\section{Limitations}
The performance of ES6D depends on the result of the 2D segmentation network \cite{xiang2017posecnn, shi2021stablepose} and the quaternion has been proved to be discontinuous in \cite{2019On}. Therefore, a unified network for instance segmentation and pose estimation, and the continuous rotation representation introduced in \cite{2019On} will be investigated in future work.

\section{Conclusion}

In this paper, a novel 6D pose estimation framework, ES6D, is proposed based on the XYZNet and A(M)GPD loss. The XYZNet is designed for feature extraction from RGB-D data. It has a fully convolutional architecture and achieves an excellent trade-off between efficiency and effectiveness. Additionally, the A(M)GPD loss is proposed to handle symmetric objects, and performs better than ADD(S) loss. Moreover, a novel numerical and visualization method is introduced to check the potential incorrect suboptimal in the loss surface.

\textbf{Acknowledgments} This work is supported by Key-Area Research and Development Program of Guangdong Province (2019B010155003) and Shenzhen Science and Technology Innovation Commission (JCYJ20200109114835623).

{\small
\bibliographystyle{ieee_fullname}
\bibliography{egbib}
}

\end{document}


\title{Supplementary material of ES6D}

\author{\textbf{Ningkai Mo} ^{1} \footnotemark[1] \\
\and
\textbf{Wanshui Gan} ^{1, 2} \footnotemark[1] \\
\and
\textbf{Naoto Yokoya} ^{2, 3} \\
\and
\textbf{Shifeng Chen} ^{1} \footnotemark[2] \\
\and
\begin{center}
 $^1$ ShenZhen Key Lab of Computer Vision and Pattern Recognition, \\
\end{center}
\and
\begin{center}
 Shenzhen Institute of Advanced Technology, Chinese Academy of Sciences, \\
\end{center}
\and
\begin{center}
$^2$ The University of Tokyo, $^3$RIKEN\\
\end{center}
\and
\begin{center}
\rm [nk.mo19941001, wanshuigan]@gmail.com, yokoya@k.u-tokyo.ac.jp, shifeng.chen@siat.ac.cn
\end{center}
}

\maketitle
\renewcommand{\thefootnote}{\fnsymbol{footnote}} 
\footnotetext[1]{The first two authors contributed equally and should be regarded as co-first authors.}
\footnotetext[2]{Corresponding author.}

\begin{figure*}
    \centering
    \includegraphics[width=1\textwidth]{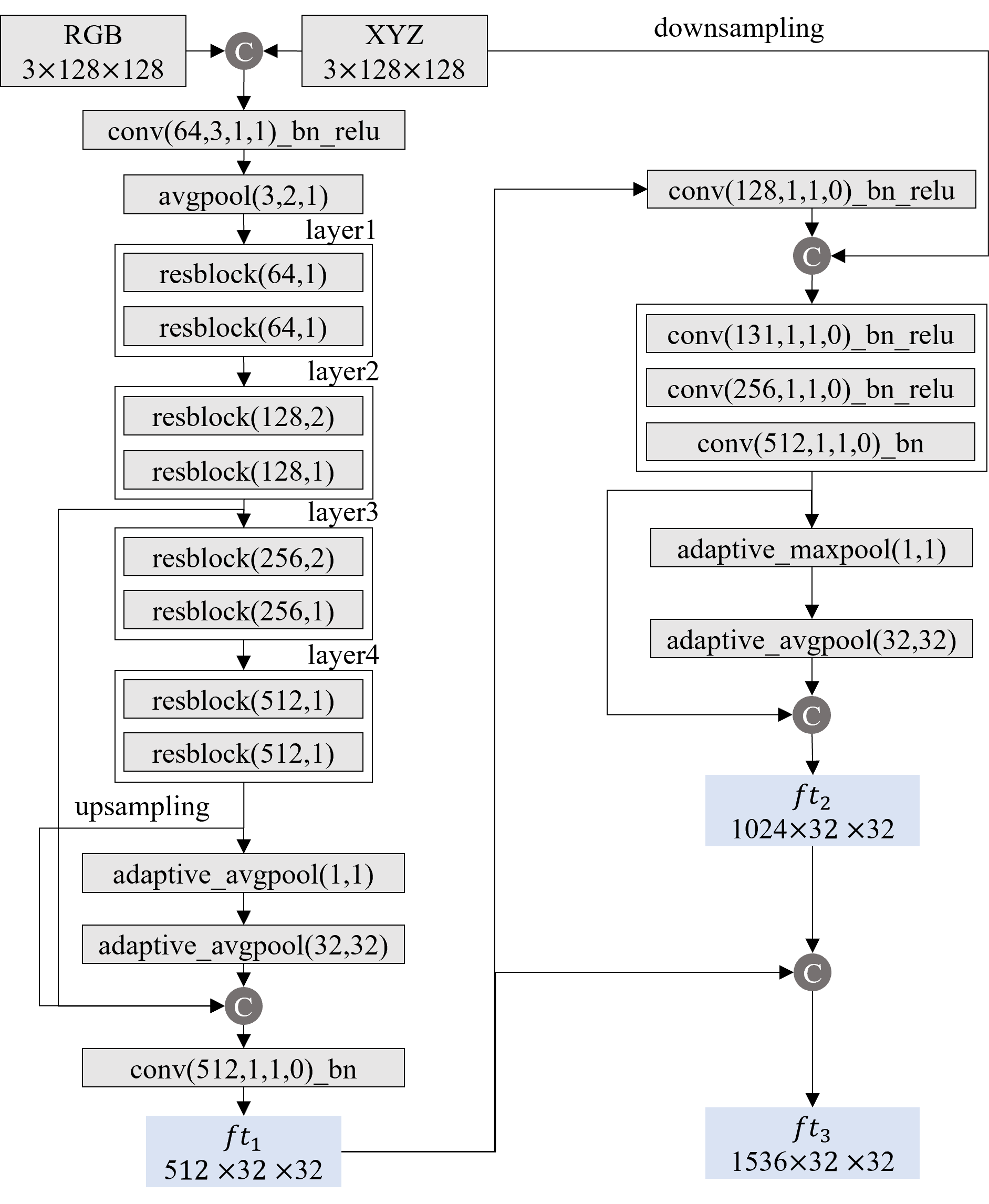}
    \caption{\textbf{The details of XYZNet}.}
    \label{fig:XYZNet}
\end{figure*}

\textbf{The details of XYZNet} As shown in Figure \ref{fig:XYZNet}, the RGB patch and the XYZ map are concatenated as the input of XYZNet. $conv(c,k,s,p)$ means the 2D convolution with output channels $c$, kernel size $k$, stride $s$ and padding $p$. $resblock(c,s)$ is the standard residual block of ResNet with output channels $c$ and stride $s$. The output of $adaptive\_avgpool(h,w)$ or $adaptive\_maxpool(h,w)$ is of size $h*w$ for any input size and the number of output features is equal to the number of input planes.

\begin{figure*}
    \centering
    \includegraphics[width=1\textwidth]{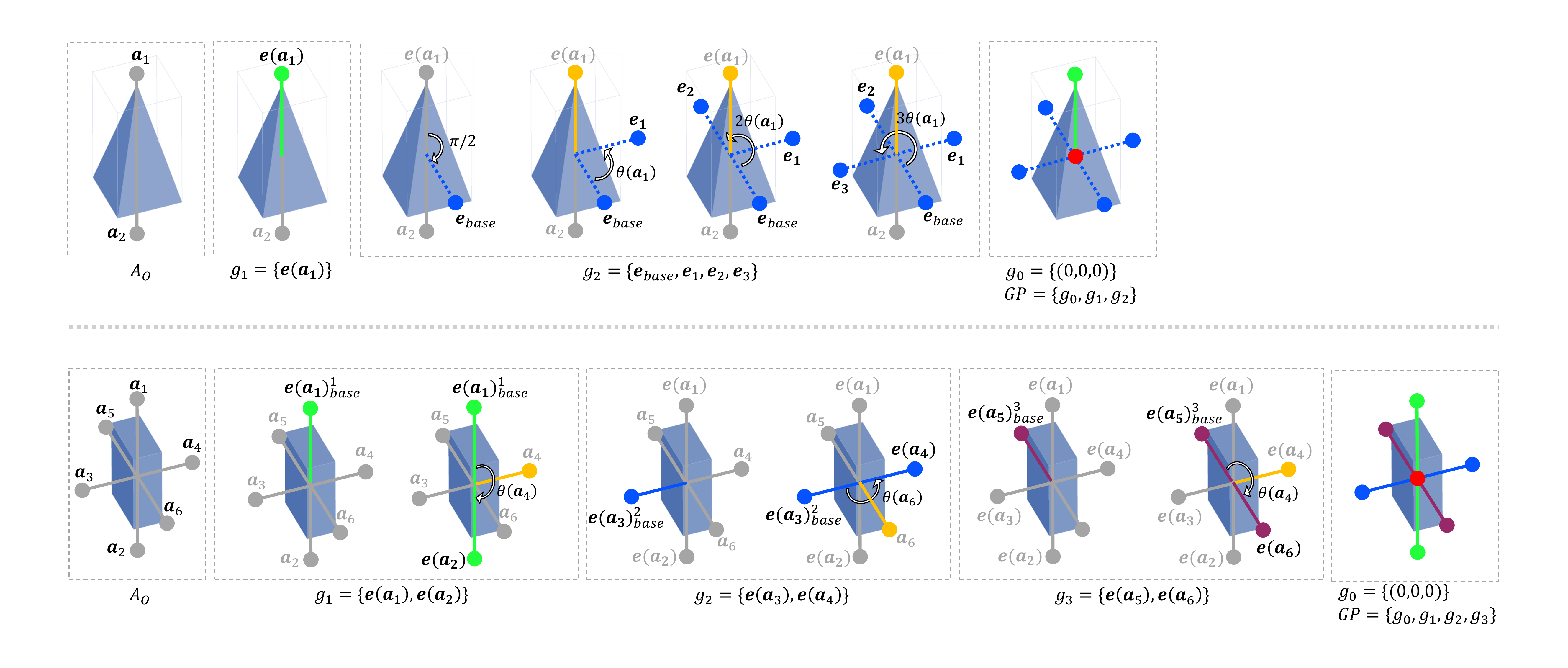}
    \caption{\textbf{Demonstration of grouping}. The first row shows the grouping operation for category 2, and the second row is for category 5.}
    \label{fig:Demonstration of grouping}
\end{figure*}

\textbf{The rules of grouping} All the categories are illustrated in the main paper (section 3.4) with the toy model. The detailed grouping principle is as follows. All the symbols are defined as the same as the main paper. 
\begin{itemize}
\item For all categories, the object centroid is the group $g_{0}$:
\begin{equation}
g_{0}=\{(0,0,0)\}.
\end{equation}

\item For category 1, there are two continuous symmetric axis-angles, we can select one as group $g_{1}$:
\begin{equation}
g_{1}=\{\boldsymbol{e}(\boldsymbol{a}_{1})\}, \quad \boldsymbol{a}_{1} \in AC_{O}.
\end{equation}

\item For category 2, there are two discrete symmetric axis-angles, we can select one as group $g_{1}$, but one axis is not enough to provide constraints for pose estimation, so we need to generate axes of number $\left|\boldsymbol{a}_{\boldsymbol{1}}\right|$ as an auxiliary group  $g_{2}$:
\begin{equation}
\begin{aligned}
g_{1} &=\{\boldsymbol{e}(\boldsymbol{a}_{1})\}, \quad \boldsymbol{a}_{1} \in A_{O};\\
g_{2} &=\left\{\boldsymbol{e}_{\text {base }}\right\} \cup\left\{R\left(\boldsymbol{a}_{1}\right)^{n} \boldsymbol{e}_{\text {base }}\right\}_{n=1}^{\left|\boldsymbol{a}_{\boldsymbol{1}}\right|-1},
\end{aligned}
\end{equation}
where $\boldsymbol{e}_{\text {base }}$ is a generated axis and is perpendicular to $\boldsymbol{e}(\boldsymbol{a}_{1})$.  The demonstration of grouping is shown in first row in Figure \ref{fig:Demonstration of grouping}.


\item For category 3, the object centroid or the $g_{0}$ is the only group.
 
\item For category 4, there are two continuous symmetric axis-angles, which make up the group $g_{1}$: 
 \begin{equation}
     g_{1}=\left\{\boldsymbol{e}\left(\boldsymbol{a}_{0}\right), \boldsymbol{e}\left(\boldsymbol{a}_{1}\right)\right\}, \quad \boldsymbol{a}_{0}, \boldsymbol{a}_{1} \in A C_{O} \wedge \boldsymbol{a}_{0} \neq \boldsymbol{a}_{1}.
 \end{equation}
 
\item For category 5, there is no continuous symmetric axis-angle. We need to group all discrete symmetric axis-angles. If the axis A could overlap with axis B after the specific angle around a symmetric axis, we regard A and B lie in the same group: 
\begin{equation}
\begin{aligned}
    g_{i}=& \{\boldsymbol{e}(\widehat{\boldsymbol{a}})_{\text {base }}^{i}\} \cup \\
    & \{ \boldsymbol{e}(\boldsymbol{a}) | 
    \boldsymbol{e}(\boldsymbol{a}) = R(\boldsymbol{\dot{ a }})^{n} \boldsymbol{e}(\widehat{\boldsymbol{a}})_{\text {base }}^{i} \\
    &  \exists \dot{\boldsymbol{a}} \in A_{o} \wedge \exists n \in [1, \cdots ,|\boldsymbol{\dot{a}}|-1] \},
\end{aligned}
\end{equation}
where $\boldsymbol{e}(\widehat{\boldsymbol{a}})_{\text {base }}^{i}$ is selected as the base of group $g_{i}$ from the ungrouped axes of $A_{O}$. The demonstration of grouping is in the second row in Figure \ref{fig:Demonstration of grouping}.

\item For the asymmetry object: 
\begin{equation}
\begin{split}
    g_{0}&=\{(0,0,0)\},\quad g_{1}=\{(1,0,0)\},\\
    g_{2}&=\{(0,1,0)\},\quad g_{3}=\{(0,0,1)\}.
\end{split}
\end{equation}
\end{itemize}

\begin{figure*}
    \centering
    \includegraphics[width=1\textwidth]{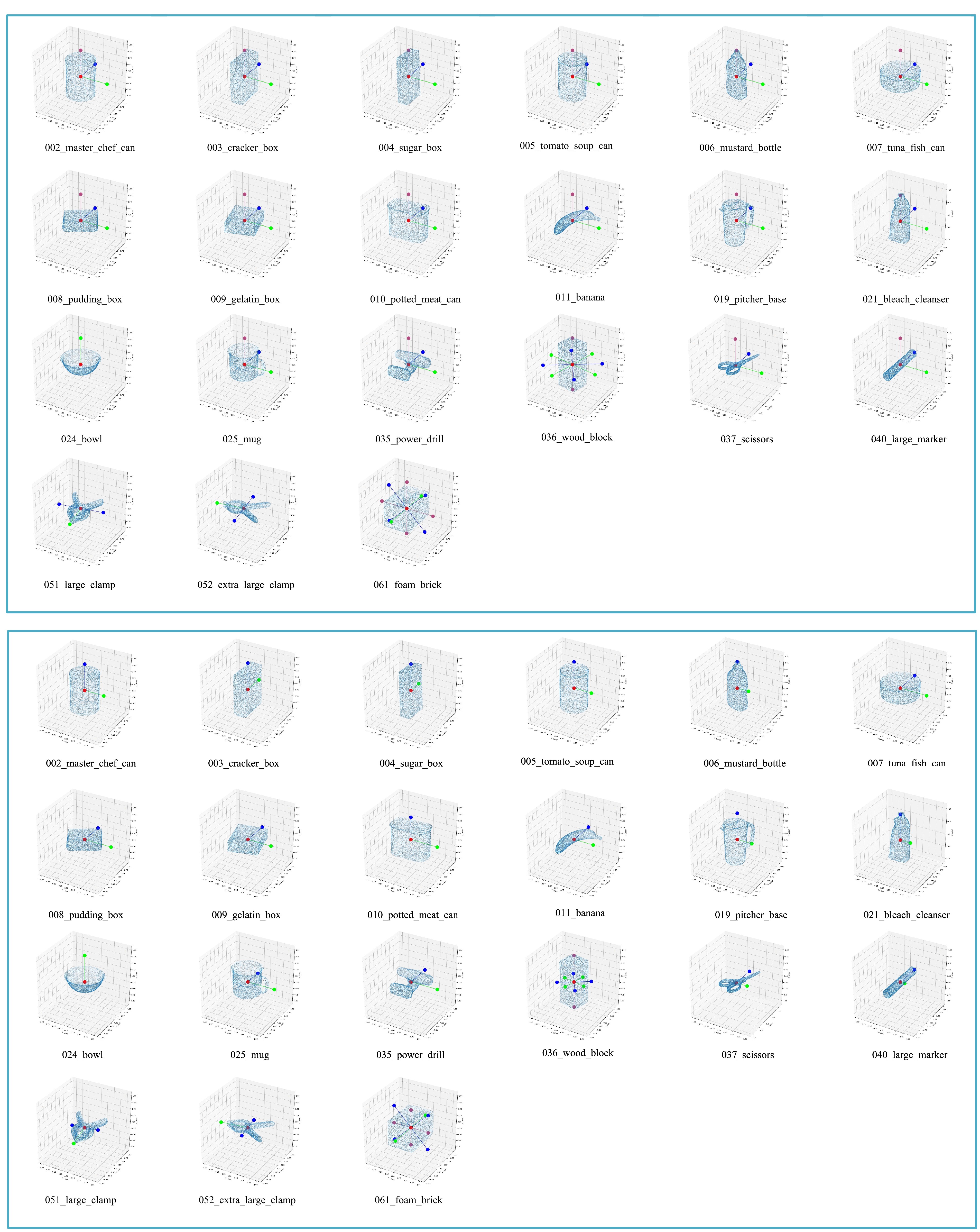}
    \caption{\textbf{The details of grouped primitives in YCB-Video dataset}. The first plot is the raw GP of objects, and the second plot is the processed GP of objects.}
    \label{fig:ycb1}
\end{figure*}

\begin{figure*}
    \centering
    \includegraphics[width=0.85\textwidth, height=1.2\textwidth]{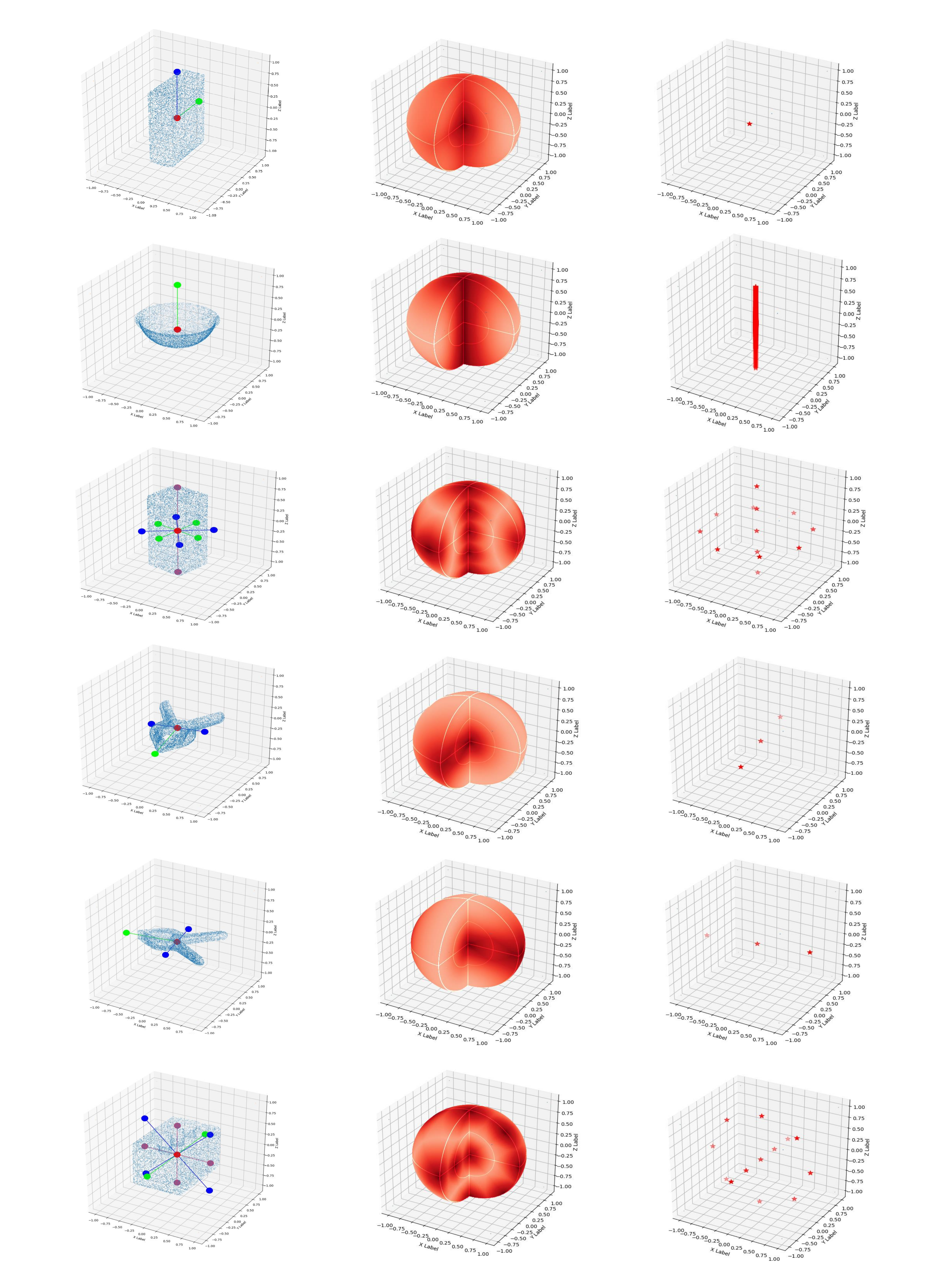}
    \caption{\textbf{The validation of processed grouped primitives in YCB-Video dataset}.  For each object, the first column presents the grouped primitives. The second shows the  A(M)GPD landscape in the rotation space, where the darker color represents the smaller value of A(M)GPD. The third column reveals the minima in each landscape. Best viewed in color.}
    \label{fig:ycb2}
\end{figure*}

\textbf{The details of grouped primitives in YCB-Video dataset \cite{xiang2017posecnn}} As shown in Figure \ref{fig:ycb1}, the first plot is the raw GP of objects in YCB-Video. For
better performance, the GP is scaled by the object's shape and we trim the GP of the asymmetry object by the principal component of the object's shape, as shown in the second plot of Figure \ref{fig:ycb1}. We also check the validation of these processed GP by the numerical and visualization method, the results are illustrated in Figure \ref{fig:ycb2}.

\begin{figure*}
    \centering
    \includegraphics[width=0.87\textwidth, height=1.2\textwidth]{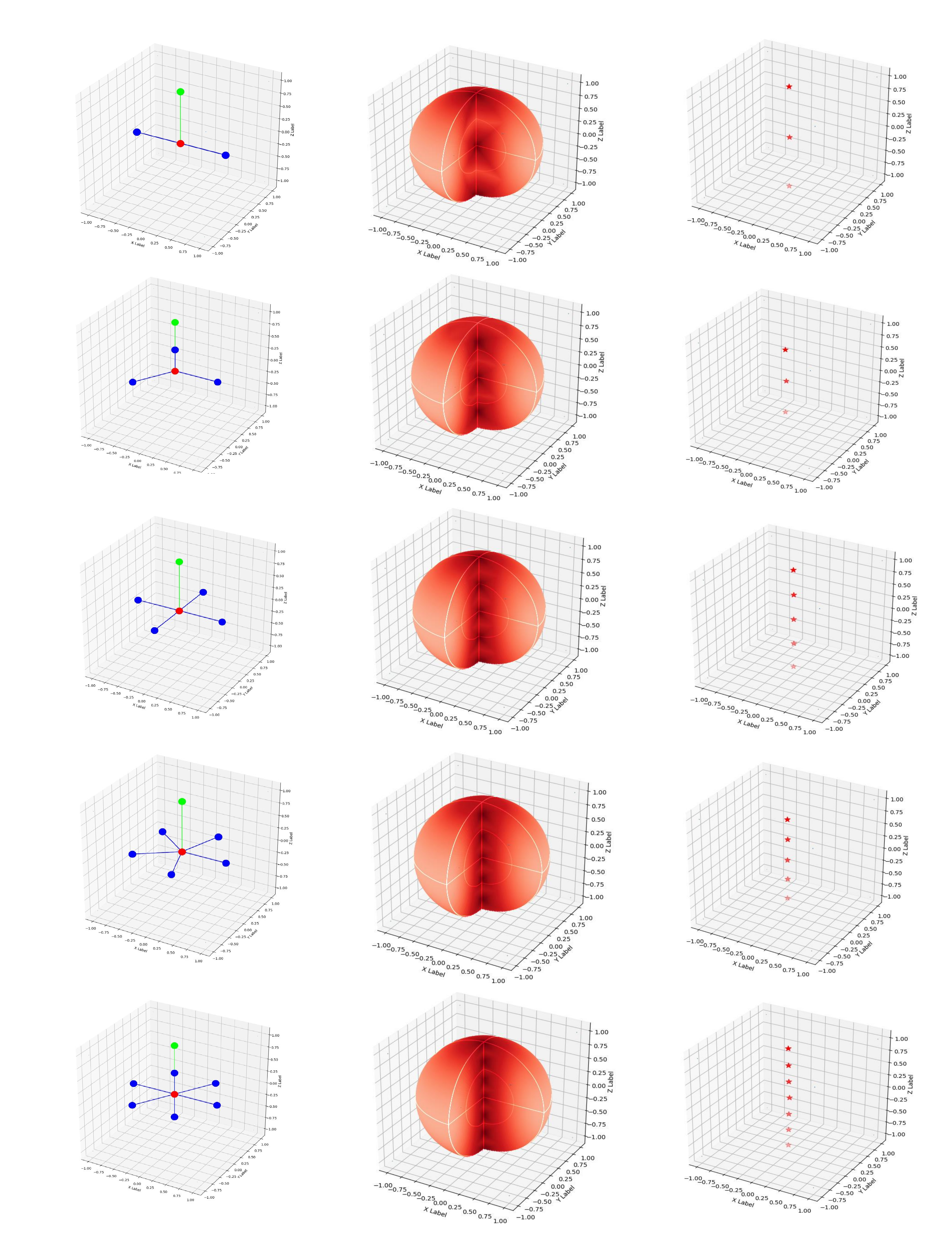}
    \caption{\textbf{More instances of category 2}. For each example, the first column presents the grouped primitives. The second shows the  A(M)GPD landscape in the rotation space, where the darker color represents the smaller value of A(M)GPD. The third column reveals the minima in each landscape. Best viewed in color.}
    \label{fig:C2}
\end{figure*}

\begin{figure*}
    \centering
    \includegraphics[width=0.8\textwidth,  height=0.48\textwidth]{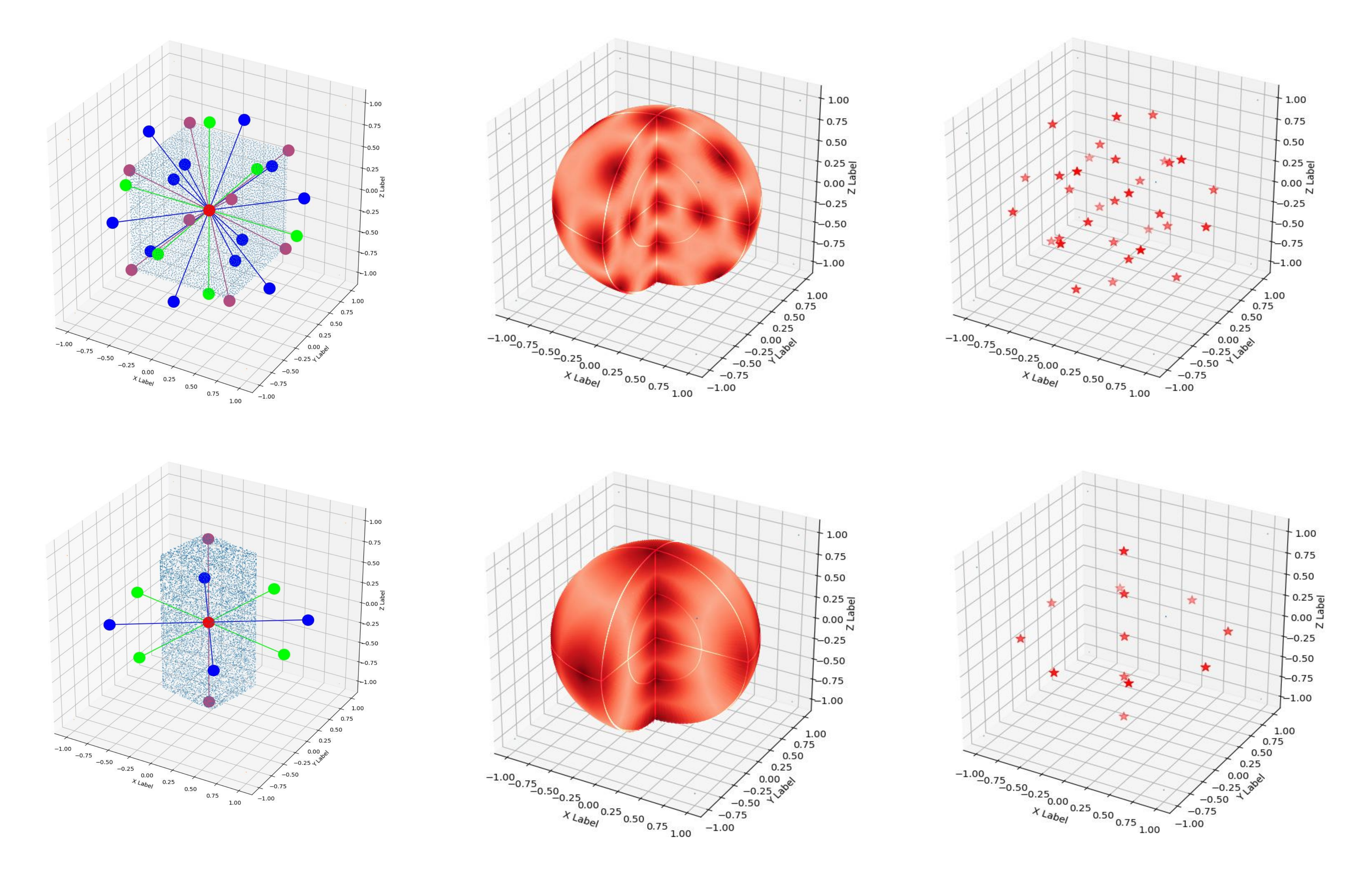}
    \caption{\textbf{More instances of category 5}. For each example, the first column presents the grouped primitives. The second shows the  A(M)GPD landscape in the rotation space, where the darker color represents the smaller value of A(M)GPD. The third column reveals the minima in each landscape. Best viewed in color.}
    \label{fig:C5}
\end{figure*}

\textbf{More instances of category 2 and 5 in symmetry} As stated in our main paper, symmetry objects can be divided into five categories. There is only one instance per category except for category 2 and 5, so we display more instances of category 2 in Figure \ref{fig:C2} and more instances of category 5 in Figure \ref{fig:C5}. The first instance in Figure \ref{fig:C5} is a regular hexahedron, which has $6+12+8$ symmetry axes, and there are $2*6+1*12+1*8+1$ minima in the A(M)GPD landscape.

\textbf{Complete experiment results} We give the complete experiment results of YCB-Video dataset in Table \ref{t:YCB-Video dataset}. Besides, we also report the AUC of A(M)GPD in Table \ref{t:AMGPD_YCB-Video} for a more comprehensive evaluation. The ground truth masks are used as input in all methods in Table \ref{t:AMGPD_YCB-Video}. As Table \ref{t:AMGPD_YCB-Video} illustrated, the proposed ES6D shows that training with the A(M)GPD loss can bring excellent accuracy on both symmetry and asymmetry objects.

\begin{figure*}
    \centering
    \includegraphics[width=1\textwidth]{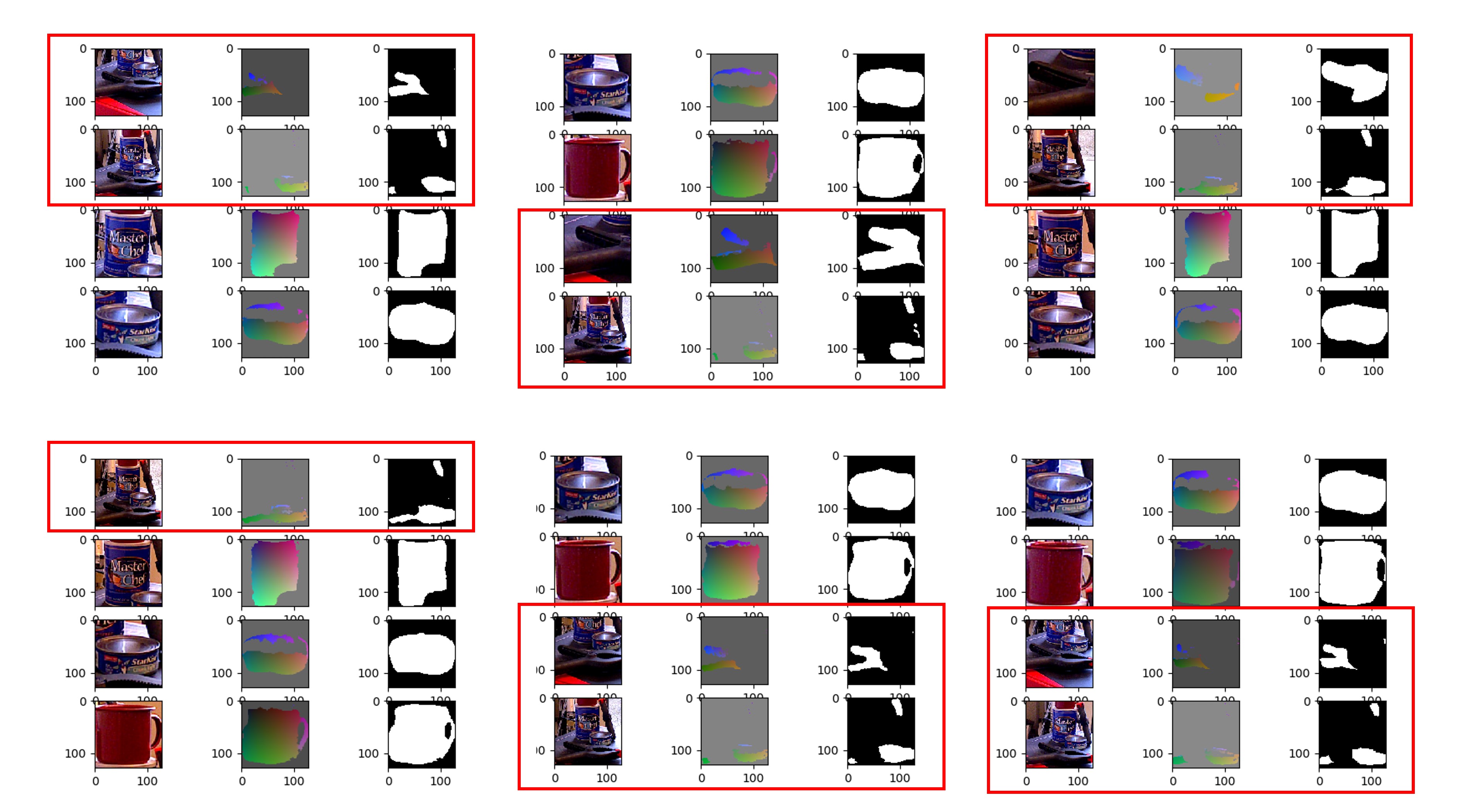}
    \caption{\textbf{Visualization for \textit{051\_large\_clamp} and \textit{052\_extra\_large\_clamp} on the YCB-Video testing dataset}. The \textit{051\_large\_clamp} and \textit{052\_extra\_large\_clamp} are marked with the red rectangle. The mask result comes from \cite{xiang2017posecnn}.}
    \label{fig:3333}
\end{figure*}

\textbf{Observation on two bad cases} In the experiment with PoseCNN segment mask and ground truth mask, we can see that there is a large gap between the symmetric objects, \textit{051\_large\_clamp} and \textit{052\_extra\_large\_clamp}, as listed in Table \ref{t:YCB-Video dataset}. We further investigate the reason for this phenomenon. We visualize the mask of \textit{051\_large\_clamp} and \textit{052\_extra\_large\_clamp} in Figure \ref{fig:3333}. We can see that, semantic segmentation network failed in distinguishing two objects with the same appearance, this would raise the category level mistake and further lead to the low accuracy on pose estimation. Furthermore, one mask in semantic segmentation could include both \textit{051\_large\_clamp} and \textit{052\_extra\_large\_clamp} part, since we crop the RGB image based on the mask result, the wrong mask prediction would lead to the extra cropped region. This would lead to a larger domain gap between the training dataset and the testing dataset. We conclude the above two reasons resulting in these two bad cases. On the other hand, in the experiment with the ground truth mask, ES6D achieves the best performance compared with PVN3D \cite{he2020pvn3d}.

\begin{figure*}
    \centering
    \includegraphics[width=0.85\textwidth, height=1.2\textwidth]{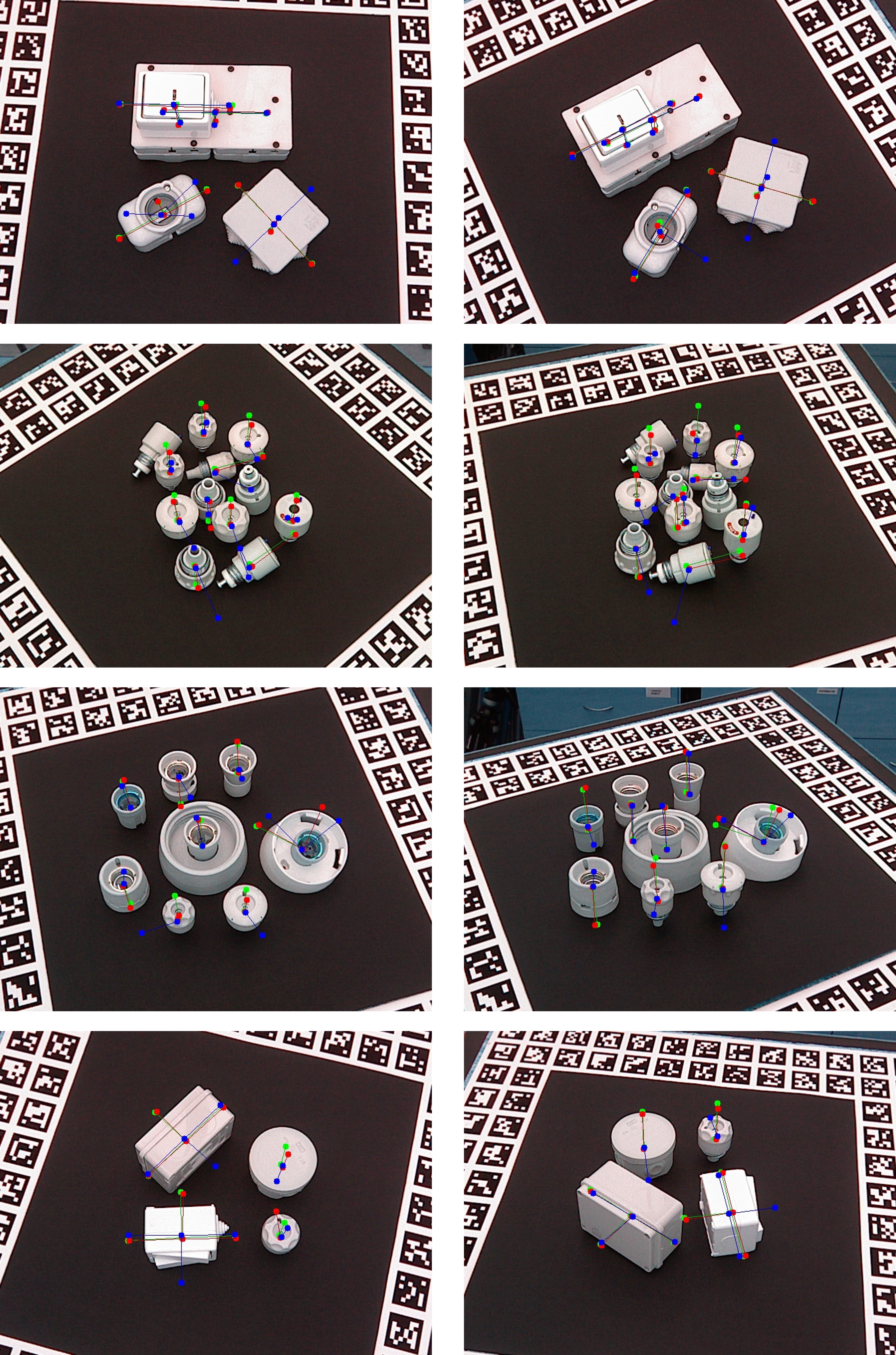}
    \caption{\textbf{Visualization on the T-LESS dataset with different training loss}. The green, red, and blue lines represent the ground truth pose, the result from A(M)GPD loss, and the result from ADD(S) loss, respectively. }
    \label{fig:1111}
\end{figure*}

\begin{figure*}
    \centering
    \includegraphics[width=0.85\textwidth, height=1.2\textwidth]{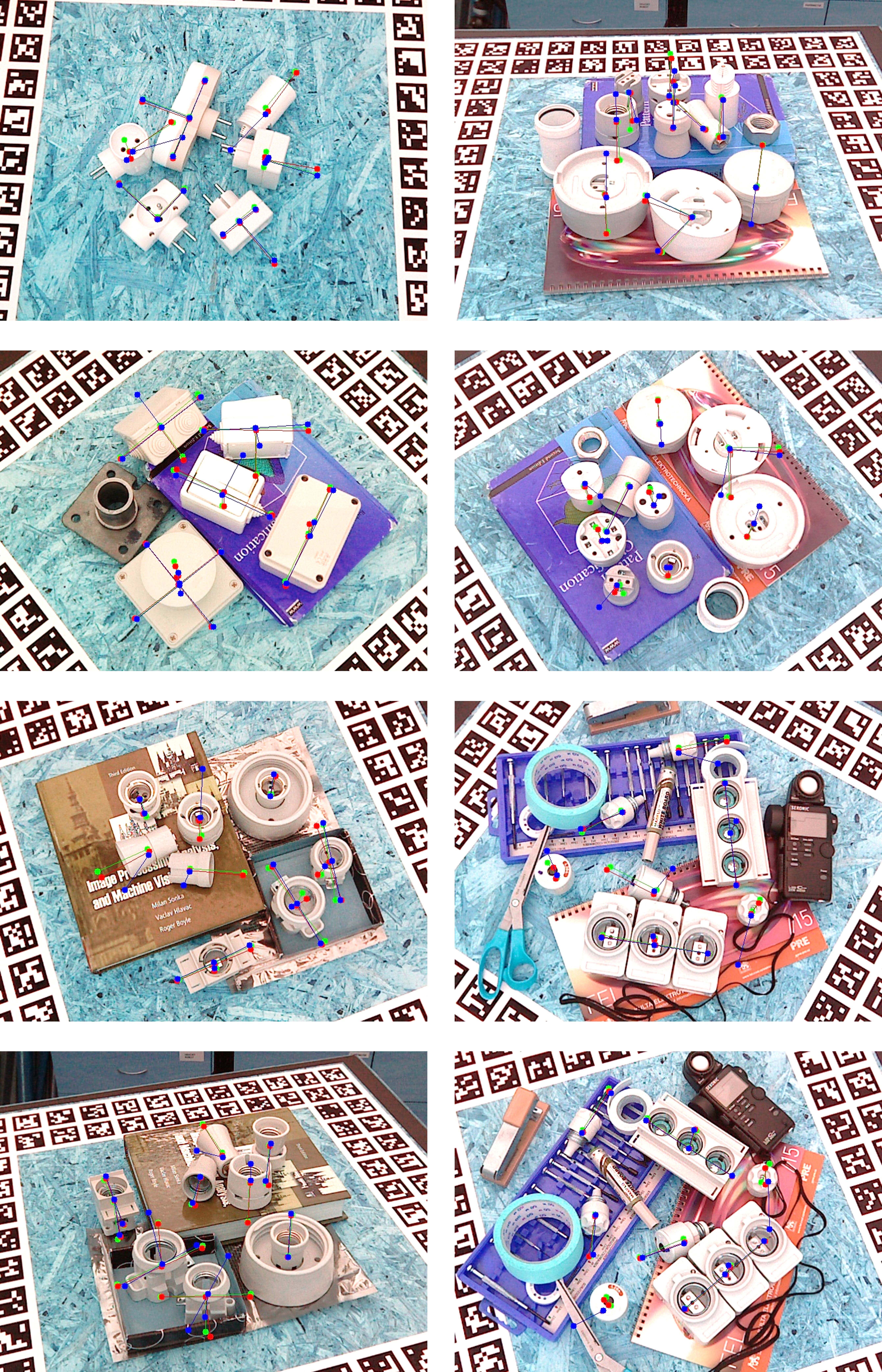}
    \caption{\textbf{Visualization on the T-LESS dataset with different training loss}. The green, red, and blue lines represent the ground truth pose, the result from A(M)GPD loss, and the result from ADD(S) loss, respectively.}
    \label{fig:2222}
\end{figure*}

\textbf{More visualization result on T-LESS dataset \cite{TLESS}} We provide more visualization results in Figure \ref{fig:1111} and Figure \ref{fig:2222} for comparison. The green, red, and blue lines represent the ground truth pose, the result from A(M)GPD loss, and the result from ADD(S) loss, respectively. We can observe that the result from A(M)GPD loss is more accurate than the result from ADD(S) loss. The result from ADD(S) loss sometimes could be totally reversed due to the local minimal problem. The proposed A(M)GPD loss could effectively address this problem.

\textbf{Comparison in YCB-Video dataset \cite{xiang2017posecnn} with video demonstration} For a visualization comparison with \cite{he2020pvn3d} and \cite{wang2019densefusion}, we make the demonstration video as attached for reference. We can see that our result is more accurate and stable compared with \cite{he2020pvn3d} and \cite{wang2019densefusion}.

\begin{table*}[t]
\centering
\scalebox{1.0}{
\begin{tabular}{l|c|c|c|c}

\hline
&  DenseFusion (iterative) & PVN3D & XYZNet with ADD(S) loss &  ES6D

\tabularnewline
\hline

002\_master\_chef\_can & 82.9 &	86.1 &	85.1 &	84.8

\tabularnewline

003\_cracker\_box & 92.5 &	95.0 &	96.1 &	96.2

\tabularnewline

004\_sugar\_box & 97.6	& 96.3 & 98.2 &	98.2

\tabularnewline

005\_tomato\_soup\_can & 90.7 &	90.9 &	93.5 &	94.0

\tabularnewline

006\_mustard\_bottle & 96.9 &	96.6 &	98.4 &	98.6

\tabularnewline

007\_tuna\_fish\_can & 75.5 &	87.0 &	92.6 &	91.9

\tabularnewline

008\_pudding\_box & 96.8 &	95.4 &	97.1 &	97.4

\tabularnewline

009\_gelatin\_box & 98.2 &	95.6 &	98.5 &	98.7

\tabularnewline

010\_potted\_meat\_can & 86.8 &	88.3 &	89.1 &	88.2

\tabularnewline

011\_banana & 77.2 & 91.9 &	94.6 &	96.3

\tabularnewline

019\_pitcher\_base & 97.4 &	96.7 &	97.9 &	98.0

\tabularnewline

021\_bleach\_cleanser & 95.3 &	93.6 &	94.0 &	95.6

\tabularnewline

\textbf{024\_bowl} & 52.8 &	68.7 &	42.9  &	94.2

\tabularnewline

025\_mug & 94.1 &	94.6 &	92.9 &	93.0

\tabularnewline

035\_power\_drill & 95.9 &	95.0 &	97.0 &	97.2

\tabularnewline

\textbf{036\_wood\_block} & 94.0 &	80.2 &	91.8 &	92.7

\tabularnewline

037\_scissors & 81.9 &	92.6 &	80.4 &	65.2

\tabularnewline

040\_large\_marker & 95.5 &	92.6 &	94.1 &	94.1

\tabularnewline

\textbf{051\_large\_clamp} & 38.7 &	79.8 &	90.9	& 92.9

\tabularnewline

\textbf{052\_extra\_large\_clamp} & 36.8 &	59.1 &	90.4&	92.6

\tabularnewline

\textbf{061\_foam\_brick} & 63.1 & 91.7 & 94.6  & 92.7

\tabularnewline
\hline

ALL &  83.7 &	89.1 &	91.9 &	\textbf{93.5}

\tabularnewline
\hline

\end{tabular}
}
\caption{The comparison of 6D Pose (A(M)GPD)) on YCB-Video Dataset. Object names with bold typeface are symmetric.}
\label{t:AMGPD_YCB-Video}
\end{table*}

\begin{table*}[t]
\centering
\scalebox{0.78}{
\begin{tabular}{l|c|c|c|c|c|c|c|c|c|c|c|c}

\hline
 & \multicolumn{2}{c|}{} & \multicolumn{6}{c|}{With PoseCNN segment mask} & \multicolumn{4}{c}{With GT segment mask}
 \tabularnewline

\hline
&  \multicolumn{2}{c|}{FFB6D}&  \multicolumn{2}{c|}{DenseFusion  }&  \multicolumn{2}{c|}{DenseFusion } &  \multicolumn{2}{c|}{\multirow{2}*{Ours}} &   \multicolumn{2}{c|}{PVN3D \cite{he2020pvn3d}} &  \multicolumn{2}{c}{\multirow{2}*{Ours}} \\

&  \multicolumn{2}{c|}{\cite{FFF6D}}&  \multicolumn{2}{c|}{\cite{wang2019densefusion} (per-pixel)}&  \multicolumn{2}{c|}{\cite{wang2019densefusion} (iterative)} &  \multicolumn{2}{c|}{~}&   \multicolumn{2}{c|}{(post process)} &  \multicolumn{2}{c}{~} 
\tabularnewline

\hline

&ADD-S &ADD(S)&ADD-S &ADD(S) &ADD-S &ADD(S) &ADD-S &ADD(S) &ADD-S &ADD(S) &ADD-S & ADD(S) 
\tabularnewline
\hline

002\_master\_chef\_can & 96.3 & 80.6 & 95.3 & 70.7 & 96.4 & 73.2 & 96.5 & 73.0 & 96.2 & 79.2  & 97.0 & 74.8

\tabularnewline

003\_cracker\_box & 96.3 & 94.6 & 92.5 &86.9 &95.8 &94.1 & 95.3 & 94.0 & 95.9 & 94.7 & 96.7 & 95.8

\tabularnewline

004\_sugar\_box & 97.6 & 96.6 & 95.1 &90.8 &97.6 &96.5  & 97.9 & 97.3 & 97.4 & 96.4 & 98.3 & 98.2  

\tabularnewline

005\_tomato\_soup\_can & 95.6 & 89.6 & 93.8 & 84.7& 94.5 & 85.5  & 97.3 & 90.4 & 96.6 & 88.5 & 97.3 & 91.9

\tabularnewline

006\_mustard\_bottle & 97.8 & 97.0  & 95.8  &90.9 &97.3 &94.7  & 98.2 & 97.9 & 97.4 &  96.3 & 98.5 & 98.4

\tabularnewline

007\_tuna\_fish\_can & 96.8 & 88.9  & 95.7 &79.6 &97.1 & 81.9  & 97.4 & 93.7 & 96.2 & 88.6 & 97.4 & 93.5

\tabularnewline

008\_pudding\_box & 97.1 & 94.6 & 94.3 & 89.3 & 96.0 & 93.3  & 96.5 & 93.4 & 96.7 & 95.2 & 97.8 & 97.3 

\tabularnewline

009\_gelatin\_box & 98.1 & 96.9  &97.2  &95.8 &98.0 &96.7  & 97.7 & 96.5 & 97.8 & 96.2 & 98.9 & 98.9 

\tabularnewline

010\_potted\_meat\_can & 94.7 & 88.1 & 89.3 &79.6 &90.7 &83.6  & 92.5 & 84.6 & 93.6 & 88.3 & 93.4 & 86.8

\tabularnewline

011\_banana & 97.2 & 94.9 & 90.0 &76.7 &96.2 &83.3  & 97.9 & 95.8 & 96.7 & 93.6 & 97.9 & 96.8 

\tabularnewline

019\_pitcher\_base & 97.6 & 96.9  & 93.6 & 87.1 & 97.5 & 96.9  & 97.8 & 97.7 & 97.1 & 96.5 & 97.9 & 97.8

\tabularnewline

021\_bleach\_cleanser & 96.8 & 94.8 & 94.4 & 87.5 & 95.9 & 89.9  & 96.3 &  92.8 & 96.1  & 93.1  & 97.0  & 94.8

\tabularnewline

\textbf{024\_bowl} & 96.3 & 96.3  & 86.0 & 86.0 & 89.5 & 89.5 & 96.4 & 96.4 & 88.7 & 88.7 & 96.8 & 96.8  

\tabularnewline

025\_mug & 97.3 & 94.2  & 95.3 & 83.8 & 96.7 & 88.9  & 97.3 & 95.0 & 97.5 & 95.5 & 97.5 & 94.5

\tabularnewline

035\_power\_drill & 97.2 & 95.9 & 92.1 & 83.7 &  96.0 & 92.7 & 97.2 & 96.3 & 96.8 & 95.3 & 97.8 & 97.4 

\tabularnewline

\textbf{036\_wood\_block} & 92.6 & 92.6 & 89.5 & 89.5 & 92.8 & 92.8 & 94.4 & 94.4 & 91.5 & 91.5 & 96.0 & 96.0

\tabularnewline

037\_scissors & 97.7 & 95.7  & 90.1 & 77.4 & 92.0  & 77.9  & 87.1 & 61.5 & 96.9 & 93.5 & 89.6 & 71.5 

\tabularnewline

040\_large\_marker &  96.6 & 89.1 & 95.1 & 89.1 & 97.6 & 93.0 & 97.8 & 90.6 & 96.7 & 91.8 & 98.3 & 90.7 

\tabularnewline

\textbf{051\_large\_clamp} & 96.8 & 96.8  & 71.5 & 71.5 & 72.5 & 72.5  & 61.0  & 61.0 & 94.4 & 94.4 & 97.5 & 97.5

\tabularnewline

\textbf{052\_extra\_large\_clamp} & 96.0 & 96.0  & 70.2 & 70.2 &  69.9 & 69.9 & 59.6 & 59.6 & 91.1 & 91.1 & 96.8 & 96.8

\tabularnewline

\textbf{061\_foam\_brick} & 97.3 & 97.3 & 92.2 & 92.2 & 92.0 & 92.0  & 96.6 & 96.6 & 96.8 & 96.8 & 96.9 & 96.9

\tabularnewline
\hline

ALL & \textbf{96.6} & \textbf{92.7} & 91.2 & 82.9 & 93.2 & 86.1  & \textbf{93.6}  &  \textbf{89.0} & 95.7 & 91.9 &  \textbf{97.1} & \textbf{93.2}

\tabularnewline
\hline

\end{tabular}
}
\caption{The comparison of 6D Pose (ADD-S, ADD(S)) on YCB-Video Dataset. Object names with bold typeface are symmetric, and the number with bold typeface means the best result.}
\label{t:YCB-Video dataset}
\end{table*}

{\small
\bibliographystyle{ieee_fullname}
\bibliography{egbib}
}